\documentclass{article}
\usepackage{amsmath}

\PassOptionsToPackage{numbers, compress}{natbib}

\usepackage[preprint]{neurips_2025}

\usepackage[utf8]{inputenc} 
\usepackage[T1]{fontenc}    
\usepackage{hyperref}       
\usepackage{url}            
\usepackage{booktabs}       
\usepackage{bbm}
\usepackage{amsfonts}       
\usepackage{nicefrac}       
\usepackage{microtype}      
\usepackage{xcolor}         
\usepackage{algorithm}
\usepackage{graphicx} 
\usepackage{algorithmic}
\usepackage{subfigure}
\usepackage{subcaption}
\DeclareMathAlphabet{\mathbbold}{U}{bbold}{m}{n}
\title{TRACE for Tracking the Emergence of Semantic Representations in Transformers}

\author{Nura Aljaafari$^{1\dagger}$,~ Danilo S. Carvalho$^{3}$,~ Andr\'{e} Freitas$^{1,2,3}$ \\
  $^{1}$ Department of Computer Science, University of Manchester, United Kingdom\\
  $^{2}$ Idiap Research Institute, Switzerland\\
  $^{3}$ National Biomarker Centre, CRUK-MI, University of Manchester, United Kingdom\\
  \texttt{\{firstname.lastname\}@[postgrad.]$^{\dagger}$manchester.ac.uk}}

\begin{document}

\maketitle

\begin{abstract}
Modern transformer models exhibit phase transitions during training, distinct shifts from memorisation to abstraction, but the mechanisms underlying these transitions remain poorly understood. Prior work has often focused on endpoint representations or isolated signals like curvature or mutual information, typically in symbolic or arithmetic domains, overlooking the emergence of linguistic structure. We introduce TRACE (Tracking Representation Abstraction and Compositional Emergence), a diagnostic framework combining geometric, informational, and linguistic signals to detect phase transitions in Transformer-based LMs. TRACE leverages a frame-semantic data generation method, ABSynth, that produces annotated synthetic corpora with controllable complexity, lexical distributions, and structural entropy, while being fully annotated with linguistic categories, enabling precise analysis of abstraction emergence. Experiments reveal that (i) phase transitions align with clear intersections between curvature collapse and dimension stabilisation; (ii) these geometric shifts coincide with emerging syntactic and semantic accuracy; (iii) abstraction patterns persist across architectural variants, with components like feedforward networks affecting optimisation stability rather than fundamentally altering trajectories. This work advances our understanding of how linguistic abstractions emerge in LMs, offering insights into model interpretability, training efficiency, and compositional generalisation that could inform more principled approaches to LM development.
\end{abstract}

\section{Introduction}\label{sec:intro}
Transformer models \footnote{Throughout this paper, we use the term "transformer" to refer to Transformer-based architectures as implemented in \citet{vaswani2017attention}} exhibit evolving internal representations during training, with recent work showing that these representations undergo \textit{phase transitions}—abrupt shifts in representational structure, generalisation behaviour, and learning dynamics~\citep{lee2024geometric, nanda2023progress, power2022grokking}. These transitions mark critical points where models reorganise internal representations and develop increasingly abstract, structured encodings~\citep{ansuini2019intrinsic, valeriani2023geometry, lee2024geometric}.

Understanding the mechanisms and timing of these shifts is essential for interpretability, model steering, and failure mode detection~\citep{grosse2024studying}. While prior studies have characterised models' behaviour at convergence or via final-layer probes, less is known about how internal linguistic structures form over the course of training. Furthermore, much of the literature on transformer interpretability and representation focuses on algorithmic or mathematical tasks~\citep{li2023systematic, zhong2024algorithmic, zhou2024what}, or examines geometric properties at the level of individual tokens or local concepts~\citep{park2024the, valeriani2023geometry}. These works leave open the question of how holistic, sentence-level semantic representations arise in transformer representations.

\begin{figure}
    \centering
    \includegraphics[width=.8\linewidth]{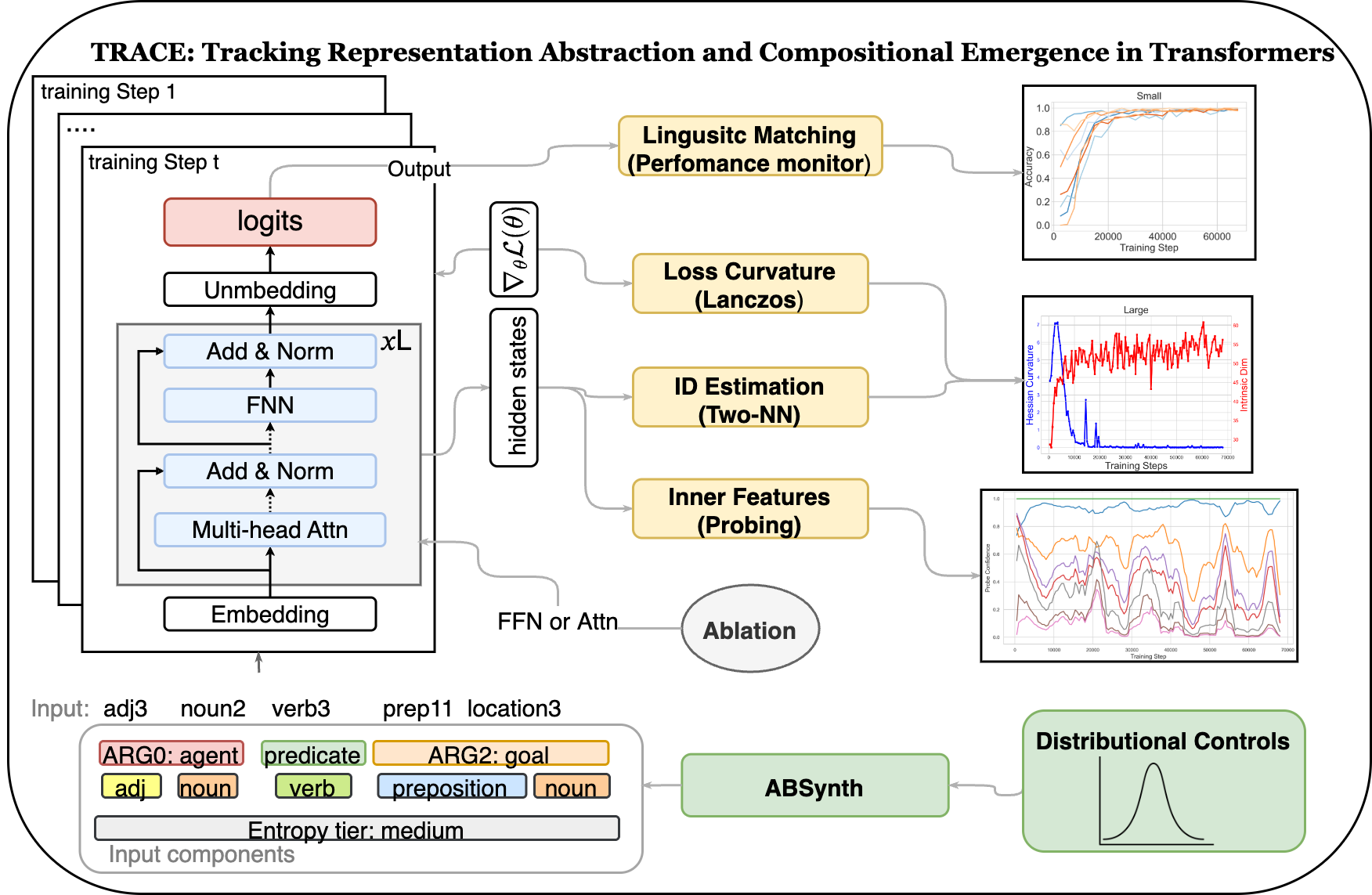}
    \caption{Overview of the TRACE framework, which integrates the monitoring of (i) intrinsic dimensionality of hidden states, (ii) spectral curvature complexity of the loss landscape, and (iii) linguistic alignment via probing and output accuracy. Inputs are sampled from ABSynth, our proposed synthetically generated corpus grounded on frame-based representations and controlled distributions over entropy, frequency, and complexity.}    \label{fig:method_overview}
\end{figure}

We address this gap by introducing \textbf{TRACE: }\emph{Tracking Representation Abstraction and Compositional Emergence} (Fig.~\ref{fig:method_overview}), a diagnostic framework that combines geometric, linguistic, and information-theoretic signals to characterise how transformers transition from memorisation to abstraction \footnote{ We use "compositional emergence" to denote the formation of structured internal representations (e.g., roles, syntactic categories), rather than formal compositional generalisation.}. 



Our central hypothesis is that abstraction emerges through a measurable phase transition, marked by: (i) characteristic rise-then-stabilise patterns in intrinsic dimensionality of hidden representations; (ii) transient spikes in loss curvature; (iii) surges in linguistic category alignment, particularly for part-of-speech and semantic accuracy; and (iv) decrease in mutual information between input and hidden representations. We test whether these phenomena, each informative in isolation, exhibit coordinated temporal dynamics that can serve as reliable sentence-level representation/abstraction indicators.

To isolate model dynamics from data confounds, we introduce a synthetic data generation framework, \textbf{ABSynth}, based on formal frame semantics~\citep{fillmore1982frame, baker-etal-1998-berkeley-framenet}. Unlike template-based approaches, ABSynth samples from abstract event frames with predefined semantic roles (agent, patient, etc.), producing corpora with transparent syntactic and semantic structure. We instantiate this framework with \textbf{ABSynth25K}, enabling precise tracking of how theoretically-motivated linguistic abstractions emerge across layers and training iterations.

This work addresses the following research questions:
\begin{itemize}
\item What geometric and statistical signals accompany the transition from memorisation to abstraction w.r.t. sentence-level representations?
    \item When do syntactic and semantic categories emerge in transformer representations over training?
    \item What mechanisms and training dynamics trigger these phase transitions, and how do architectural and optimisation factors influence their onset?
\end{itemize}

This paper makes three key contributions. First, we introduce \textbf{TRACE}, a unified diagnostic framework that jointly tracks abstraction and early representational structure formation using coordinated geometric (curvature), statistical (mutual information, intrinsic dimensionality), and linguistic (probing-based) signals throughout training. Second, \textit{an original spectral curvature complexity measure }$\mathcal{C}(H)$ characterising loss landscape properties. Third, a \textit{frame-semantics grounded synthetic sentence generation framework \textbf{ABSynth}}, from which we derive the supporting \textit{\textbf{ABSynth-25K} corpus} for controlled analysis of representational dynamics.

Across all experiments, we observe a consistent phase transition, indicated by coordinated shifts in curvature flattening, intrinsic dimensionality, and probe performance, which marks the onset of abstraction. These patterns persist across model scales and ablation variants, pointing to structural regularities in transformer learning dynamics. Understanding these transitions supports designing more interpretable, adaptive, and resource-efficient language models.

\section{Related Work}\label{sec:rl}
\paragraph{Phase Transitions in Training Dynamics.}
Representational transitions during training have been documented in small-scale settings like grokking \citep{nanda2023progress, power2022grokking}, where models shift from memorisation to generalisation. \citet{lee2024geometric} attributed these to geometric reorganisation, while \citet{clauw2024informationtheoretic} identified emergent information-theoretic structure. \citet{nakkiran2021deep} described double-descent phenomena, where generalisation is preceded by changes in spectral behaviour. Stagewise development has also been observed in attention heads and induction circuits \citep{olsson2022context}.

\paragraph{Loss Landscape Geometry and Generalisation.} Curvature properties help illuminate learning dynamics in neural networks \citep{bottcher2024visualizing, wang2024loss}. Early work linked sharp minima to overfitting \citep{keskar2016large}, while later studies found flatter regions correspond to better generalisation \citep{hao2019visualizing, sankar2021deeper}. Spectral metrics like Hessian trace and effective rank capture curvature anisotropy \citep{ahn2023escape, bottcher2024visualizing}, with recent transformer analyses showing systematic evolution of these metrics during training \citep{hoogland2024developmental, wang2024loss}. Empirically, models trained in overparameterised regimes often exhibit flat Hessian spectra with many near-zero eigenvalues \citep{singh2021analytic}, corresponding to improved generalisation, indicative of abstract representation formation \citep{ahn2023escape}. 

\paragraph{Intrinsic Dimensionality and Representation Compression.} 
Intrinsic dimensionality (ID) serves as a proxy for the representational complexity of neural networks \citep{aghajanyan2021intrinsic, facco2017estimating, ansuini2019intrinsic}. Under the manifold hypothesis, high-dimensional activations are assumed to lie on lower-dimensional submanifolds \citep{cayton2008algorithms}, with the ID reflecting the necessary degrees of freedom to explain observed variation. During training, representations typically exhibit a rise–fall pattern—initially increasing as features entangle, then compressing as abstraction emerges \citep{ansuini2019intrinsic, cheng2025emergence, valeriani2023geometry}. While \citet{aghajanyan2021intrinsic} showed pre-trained models can be fine-tuned in low-dimensional reparameterisations, \citet{cheng-etal-2023-bridging} and \citet{lee2024geometric} correlate geometric compression with linguistic information acquisition. Recent work confirms linguistic features occupy low-dimensional subspaces \citep{razzhigaev-etal-2024-shape} and that compressibility enables compositional generalisation \citep{elmoznino2024complexity}. ID estimation methods range from PCA \citep{little2017multiscale} to geometric approaches like TWO-NN \citep{facco2017estimating} and GRIDE \citep{denti2022generalized}.

\paragraph{Intermediate Layers and Representation Studies.} Recent studies highlight intermediate layers' role in shaping model representations \citep{NEURIPS2024_68716328, fan2024, skean2025layer}. These layers often show stronger linguistic alignment than final layers \citep{skean2025layer}. \citet{lepori2023break} introduced structural compositionality, revealing that models can decompose complex tasks into modular subroutines, with intermediate layers playing a crucial role in this decomposition process. 
Related work on symbolic domains examines abstraction in transformers trained on code or algorithmic tasks \citep{li2023systematic, zhou2024what, zhong2024algorithmic}, though these focus on task-specific behaviours rather than semantic abstraction in linguistic representations.

\paragraph{Synthetic datasets}
Several synthetic benchmarks have been developed to study abstraction and generalisation in neural models, though most focus on algorithmic or symbolic reasoning tasks rather than grounded linguistic structure. SCAN~\citep{lake2018generalization} tests systematic compositional skills through command-to-action mappings, while PCFG-based datasets~\citep{hupkes2020compositionality} probe models' syntactic generalisation abilities using controlled linguistic commands. Mathematical reasoning datasets~\citep{saxton2019analysing} and algorithmic tasks~\citep{power2022grokking} provide controlled environments for studying learning dynamics but lack linguistic structure.

Unlike prior work that investigates abstraction using symbolic or algorithmic datasets with limited linguistic grounding, our approach targets sentence-level semantic abstraction in transformer models trained on structured, English-like input. We introduce a synthetic corpus generator grounded in frame semantics, enabling precise control over contextual entropy, role structure, and token distributions. This design reflects key properties of natural language while retaining full annotation and sampling transparency. While previous studies provide valuable insights, they often examine a single diagnostic signal in isolation, and are typically restricted to tasks or domains that do not generalise to realistic linguistic settings or scale to larger models. By contrast, our multi-signal approach offers a holistic view of how abstraction emerges during training. This principled integration enables more transferable and interpretable analysis of representation learning in modern transformers.

\section{Methodology}\label{sec:methodolgy}
As shown in Fig.~\ref{fig:method_overview}, our method integrates the following signals to detect phase transitions during transformer training: spectral curvature of the loss landscape, intrinsic dimensionality of representations, and linguistic category alignment. Our intuition is that these metrics capture complementary aspects of representation learning: optimisation dynamics reflect updates in model weights, geometric measures and probing reveal reorganisation of semantic relationships, and linguistic alignment reflects emergent structure in the model’s outputs. We define semantic abstraction as the model's ability to internalise role-based generalisations that extend beyond surface lexical forms—for example, recognising the \texttt{ARG2} role regardless of whether it is realised as "noun3", or "location22". In this setting, abstraction is evidenced when internal representations align with underlying semantic functions rather than memorised token identities. The following sections detail each aspect, along with our synthetic data generator and experimental setup.

\subsection{Spectral Signals of Loss Landscape Geometry}
We characterise loss landscape geometry using Hessian-based curvature metrics to detect structural shifts in representation learning. We adopt a scalable approximation via the Lanczos algorithm~\citep{lanczos1950iteration}, which estimates the top \( K \ll N \) eigenvalues of the Hessian using efficient Hessian-vector products. Let \( \mathcal{L}(\theta) \) be the training loss with gradient \( g = \nabla_\theta \mathcal{L}(\theta) \) and Hessian \( H_\theta = \nabla^2_\theta \mathcal{L}(\theta) \). We compute the truncated spectrum \( \{\lambda_i\}_{i=1}^K \) where \( K \ll N \), motivated by observations that curvature information concentrates in dominant modes~\citep{sankar2021deeper}. Our spectral metrics include:

\begin{itemize}
    \item \textbf{Hessian Trace:} \( \mathrm{Tr}(H_\theta) = \sum_{i=1}^{N} \lambda_i \), quantifying overall curvature magnitude. Decreasing trace indicates flatter minima associated with improved generalisation \citep{sankar2021deeper, zhao2022penalizing, ahn2023escape}.


    \item \textbf{Entropy-Based Effective Rank:} \(
    r_{\text{eff}} = \exp\left(-\sum_i p_i \log p_i\right)\), where \( p_i = \frac{|\lambda_i|}{\sum_j |\lambda_j|}\); this Shannon entropy-based measure~\citep{roy2007effective} quantifies dominant curvature directions. Low values reflect curvature concentration (anisotropy)—abstraction; high values indicate distributed, isotropic curvature—memorisation \citep{elmoznino2024complexity}.
    
    
\end{itemize}

To unify curvature magnitude and directional concentration, we define a \textbf{curvature complexity score}:
\begin{equation}
    \mathcal{C}(H) = \frac{\mathrm{Tr}(H)}{\sqrt{r_{\text{eff}}}}.
\end{equation}

This measure increases with both the overall curvature and its spectral concentration. High \( \mathcal{C}(H) \) values correspond to sharp, anisotropic curvature, often reflecting representational reorganisation or memorisation. In contrast, low values denote flatter, more isotropic landscapes, typically aligned with abstraction and generalisation.
\subsection{Intrinsic Dimensionality }
\label{sec:track_ID}

We characterise abstraction in transformer representations through the lens of \textit{intrinsic dimensionality} (ID), motivated by the manifold hypothesis \citep{cayton2008algorithms}. Given hidden representations \( Z \in \mathbb{R}^D \), ID is defined as the minimal number of degrees of freedom required to locally parameterise the data distribution \citep{facco2017estimating}. That is, although \( Z \) may lie in a high-dimensional space, it may concentrate around a lower-dimensional manifold \( \mathcal{M} \subset \mathbb{R}^D \) of dimension \( d \ll D \). We adopt the TWO-NN estimator \citep{facco2017estimating}, a non-parametric, maximum likelihood estimator based on local geometry. Given a batch of activation vectors \( \{x_i\}_{i=1}^N \), the intrinsic dimension is estimated as:
\begin{equation}
\label{eq:id-estimator}
\text{ID} = \left( \frac{1}{N} \sum_{i=1}^{N} \log \left( \frac{r_2(x_i)}{r_1(x_i)} \right) \right)^{-1},
\end{equation}
where \( r_1(x_i) \) and \( r_2(x_i) \) denote distances to the first and second nearest neighbours of \( x_i \), respectively. This approach requires no tuning parameters and assumes only uniform local density. To provide a more holistic view, we average the ID over the model layers as \(\overline{ID}^{(t)} = \frac{1}{L} \sum_{\ell=1}^L ID_\ell^{(t)}, \), and use \(\overline{ID}^{(t)}\) in our analysis. By computing ID across training steps and network layers, we capture the dynamic evolution of representational structure. We hypothesise a characteristic trajectory aligned with other metrics: low initial ID during early training, rising ID during transition, and stabilisation or decrease as the model projects data onto semantically coherent structures—signalling abstraction.

\subsection{Linguistic Category Alignment}
We evaluate abstraction emergence through two complementary approaches: internal representation probing and output generation analysis. For each, we examine both semantic roles (e.g., AGENT, PATIENT) for event structure understanding, and part-of-speech (POS) categories for syntactic abstraction. 

\paragraph{Internal Probing.} We apply diagnostic probes \( p_c^{(\ell)} \) to hidden states at layer \( \ell \) to measure category-specific confidence scores during training. These probes quantify how well internal representations encode linguistic structures at each training step \( t \):
\begin{equation}
\text{Conf}_c^{(\ell,t)} = \frac{1}{|\mathcal{B}|} \sum_{x \in \mathcal{B}} p_c^{(\ell)}(h_{\ell}(x)),
\end{equation}
where \( h_{\ell}(x) \) is the hidden representation, \( \mathcal{B} \) is the evaluation batch size and \( p_c^{(\ell)} \) is a linear classifier trained on trained model frozen checkpoints. Probes are used to capture the evolving alignment between internal features and abstract linguistic categories. To validate that observed linguistic alignment reflects learned abstraction rather randomness, we trained probes on randomly initialised models. These probes performed at or near chance, confirming that linguistic features are not encoded prior to training. This supports the view that abstraction emerges progressively and is localised in stages as training evolves. Full results are reported in Appendix~\ref{app:probebaselines}.

\paragraph{Output Generation Analysis.} We also assess whether generated tokens respect linguistic constraints.  For each generated token \( \hat{y} \), we compute category-specific accuracy:
\begin{equation}
\text{Acc}_c^{(t)} = \frac{1}{|\mathcal{D}_c|} \sum_{(x_i, y_i) \in \mathcal{D}_c} \mathbbold{1}\left[\hat{y}_i \in \mathcal{f}(y_i)\right],
\end{equation}
where \( \mathcal{D}_c \) contains sequences with category \( c \), and \( \mathcal{f}(y_i) \) denotes the set of valid tokens for the expected category at position \( i \). This metric reveals whether abstract patterns learned internally are successfully deployed during generation. By jointly analysing internal representation alignment and output conformity alongside geometric metrics, we identify precisely when models transition from memorising token associations to acquiring structured abstractions. Divergence between internal and output measures reveals intermediate states where models have partially acquired abstract representations but cannot yet reliably deploy them in generation. The complete methodology for token categorisation, probe architecture, and training procedures is detailed in Appendix~\ref{app:probes_labels}.

\subsection{Information Compression via Mutual Information}
\label{sec:info_compression}
While we explored mutual information (MI) as a potential signal of abstraction, we found that MI estimates were highly volatile and did not consistently align with the phase transitions observed through other metrics. This behaviour likely stems from two issues: (i) MI estimation is inherently noisy in high-dimensional settings, and (ii) abstraction in transformers involves structural reorganisation rather than pure information compression. These patterns were consistent across both types of MI we measured: (i) $I(X; Z_\ell)$, the information retained about the input $X$ in the hidden states $Z_\ell = \phi_\ell(X)$; and (ii) $I(Z_\ell; Z_{\ell+1})$, the information shared between adjacent layers. Due to this instability, MI lacked the resolution to serve as a reliable diagnostic. We include full experimental results, estimation procedures, and MI trajectories in Appendix~\ref{app:mine} for completeness, but do not consider MI a core component of our results.

\subsection{Synthetic Data Generator}  
To isolate representational dynamics from confounding factors in natural data, we employ \textbf{ABSynth}, our controllable synthetic data generation framework grounded in frame semantics~\citep{fillmore1982frame, baker-etal-1998-berkeley-framenet}. ABSynth controllably builds synthetic corpora by sampling from structured sentence frame representations with predefined semantic roles (e.g., \texttt{AGENT}, \texttt{PATIENT}, \texttt{THEME}). The framework supports precise manipulation of structural properties, including vocabulary size, token frequency, syntactic/semantic complexity, and contextual entropy.

The generation pipeline consists of three modular components: (1) a lexicon module that assigns words to categories under a Zipfian frequency distribution~\citep{zipf1949human, piantadosi2014zipf} ($\alpha = 1.05$), augmented with variable-strength collocations; (2) a frame-based sentence constructor that assembles grammatically well-formed sentences across three levels of structural complexity; and (3) an entropy-aware token selector that modulates predictability by adjusting sampling probabilities for the corpus. For this study, we use ABSynth to generate \textbf{ABSynth25K}, a dataset of 25,000 sentences with complete frame-semantic annotations. Each example includes ground-truth semantic roles and syntactic categories (POS tags) derived directly from the underlying frame structure, enabling precise investigation of how linguistic abstractions emerge in neural representations. ABSynth25K follows an 80/10/10 training/validation/test split. Complete generation procedures and frame specifications are detailed in Appendix~\ref{app:synthetic-data}.

\subsection{Models Architectures and Training Setup}

\paragraph{Transformer Architectures.}
We train three decoder-only transformer models of increasing capacity. Small model (2-head, 128 FFN, 64 $d_{model}$, 1-layer), medium model (3-head, 384 FFN, 96 $d_{model}$, 2-layer) and large model (4-head, 512 FFN, 128 $d_{model}$, 3-layer). Models share the same positional encoding scheme, tokenisation, and vocabulary. A fixed sequence length (16) and batch size (128) are used across runs to standardise training dynamics. We record dense checkpoints throughout training to record our metrics. Complete training details, downstream task formalisations and model hyperparameters are reported in Appendix~\ref{app:implemenation_details}.


\paragraph{Ablation of Architectural Components.}  
To isolate the architectural factors driving abstraction, we ablate key transformer components by removing feed-forward (FFN) blocks and reducing the number of attention heads. These interventions are designed to test whether the capacity for abstraction depends on transformation depth or attention expressivity, and to determine which mechanisms are necessary for triggering representational phase transitions.

\section{Results \& Analysis}\label{sec:results}
\subsection{Coordinated Phase Shift in Training Dynamics}
\begin{figure}[]
    \centering
    \includegraphics[width=.8\textwidth]{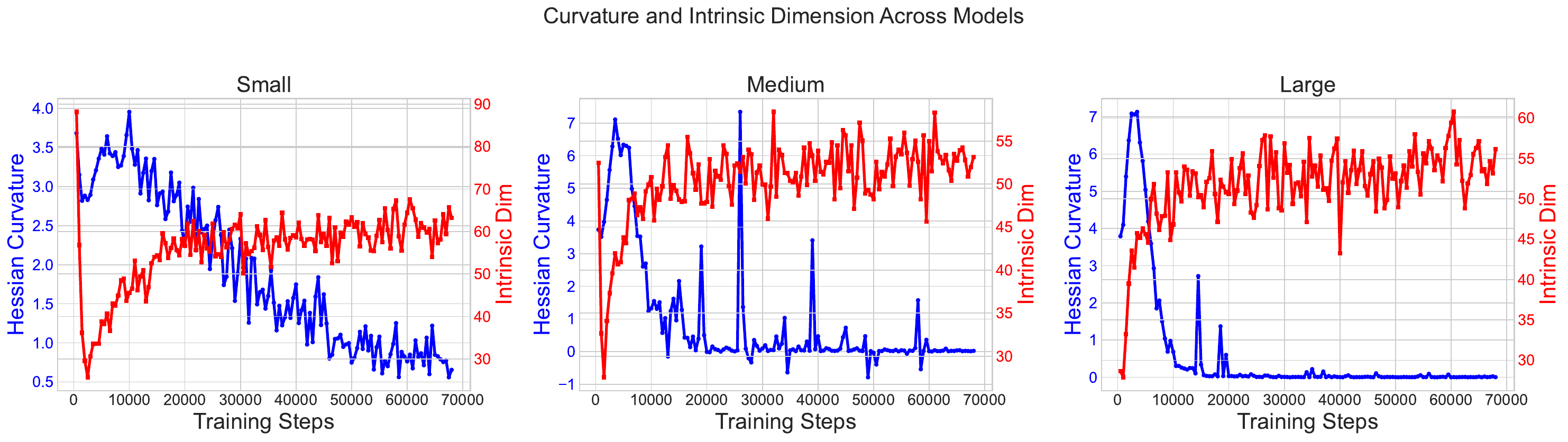}
    \\
    \includegraphics[width=.8\textwidth]{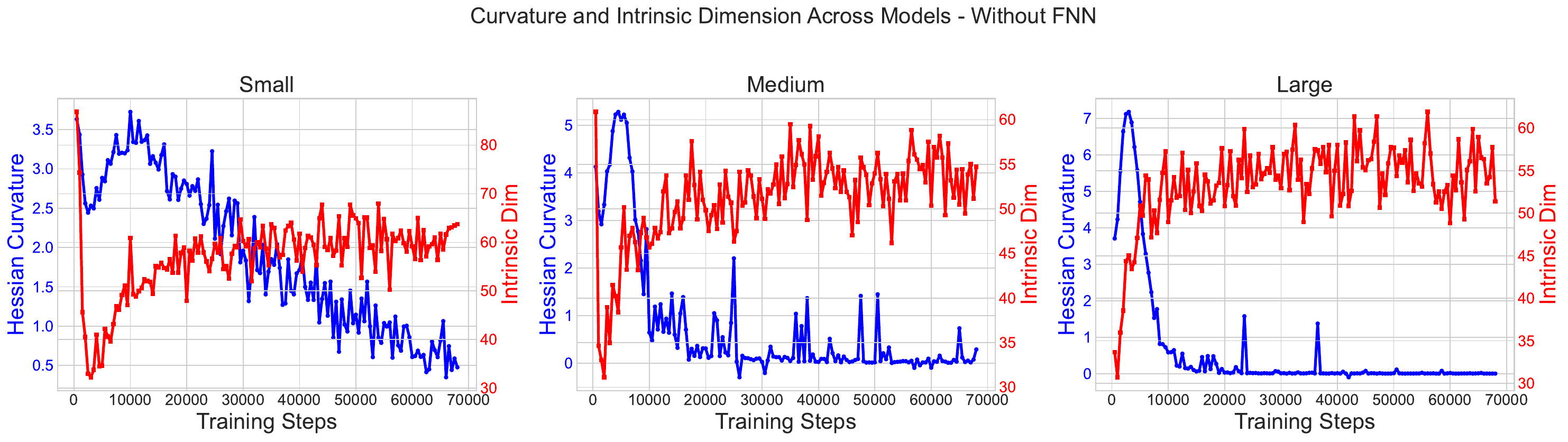}
    \\
    \includegraphics[width=.8\textwidth]{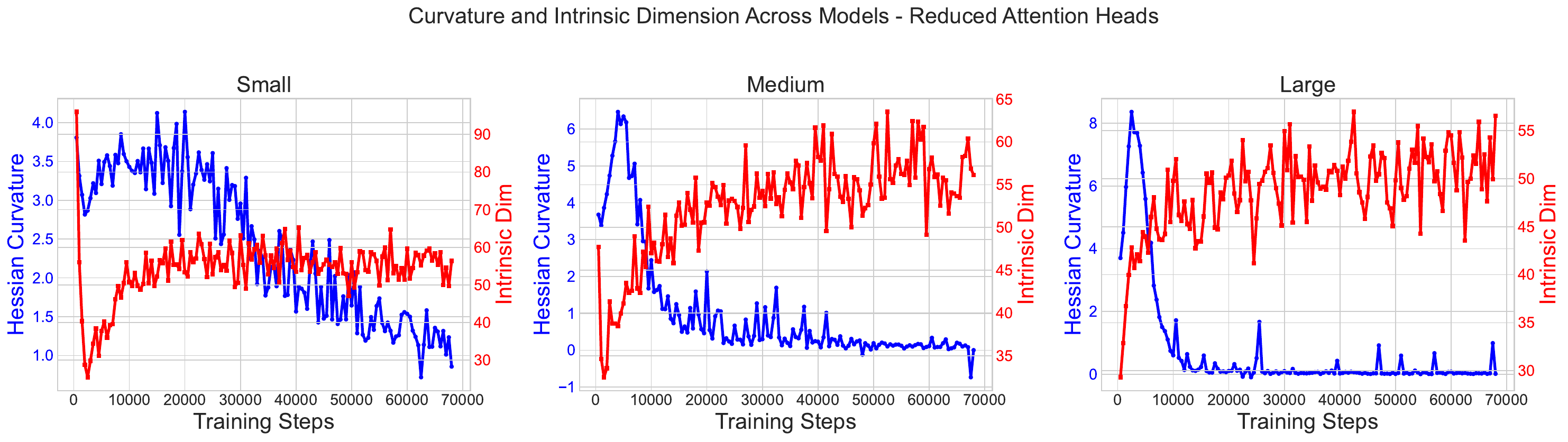}
    \caption{Coordinated dynamics of Hessian Curvature Score (blue) and Average Intrinsic Dimension (red) across training steps for three model architectures.
    Each row shows a different architectural variant: standard models (top), models without feed-forward networks (middle), and models with a single attention head (bottom). The intersection points between curvature and ID trajectories mark critical phase transitions in representational learning, with timing and stability varying across architectures but preserving the fundamental pattern.}
    \label{fig:curvature-id-comparison}
\end{figure}

Across all model configurations, we observe a robust two-phase training dynamic: an initial regime of rising ID and elevated curvature, followed by a transition into flatter curvature and stabilised representational complexity (Fig. \ref{fig:curvature-id-comparison}). This transition is marked by a consistent intersection between the Hessian curvature score (blue) and ID trajectories (red), which we interpret as a phase shift in learning dynamics.

In larger models (right panel), this intersection occurs early (around step 5000) and sharply, with curvature rapidly collapsing while ID plateaus at higher values, indicating the emergence of high-capacity, structured representations. Medium-scale models (centre panel) follow qualitatively similar transitions but exhibit notable periodic spikes in curvature throughout training. These persistent oscillations suggest recurring reorganisation events where the model temporarily revisits higher-curvature regions of the loss landscape, reflecting optimisation instabilities with limited architectural capacity.

Smaller models (left panel) exhibit delayed transitions (after step 30000), noisier trajectories, and lower equilibrium ID values, \textit{demonstrating clear architectural bottlenecks in abstraction capacity}. Despite variations in timing, amplitude, and stability, the fundamental pattern of curvature-ID intersection remains consistent, \textit{suggesting a universal geometric signature of abstraction emergence that scales with model capacity but preserves its essential character.}

\subsection{Differential Impact of Architectural Components}
Our ablation experiments examine how specific architectural components influence abstraction dynamics, revealing subtle but informative effects. Fig. \ref{fig:curvature-id-comparison} presents curvature and ID trajectories across three architectural variants: standard models (top row), models without feed-forward networks (FFNs) (middle row), and models with a single attention head (bottom row).

All variants preserve the fundamental pattern of an initial curvature peak followed by a decline, concurrent with rising ID that eventually stabilises. This consistency demonstrates that abstraction emergence is surprisingly \textit{robust to architectural modifications and may be an inherent property of transformer-based optimisation.}

Removing FFNs (middle row) increases curvature volatility, especially in small and medium models, with persistent oscillations but reduced spike amplitudes. This suggests FFNs contribute to optimisation stability and smooth representational development, functioning as distributed lookup structures \citep{dai2021knowledge}. Despite volatility, the phase transition timeline remains preserved, \textit{indicating FFNs enhance rather than enable abstraction emergence.}

Single attention head models (bottom row) show scale-dependent effects. Medium and large models exhibit more frequent but lower-amplitude spikes, \textit{revealing a stability-smoothness trade-off}. Small models show greater impact: reduced ID values and delayed phase transitions, indicating \textit{attention capacity constraints affect smaller architectures more severely}. Nevertheless, the fundamental curvature-ID intersection pattern persists across all configurations.

These results demonstrate the architectural resilience of core learning dynamics in transformers. The preserved geometric signature across configurations establishes that abstraction is observable and a \textit{fundamental property of overall transformers' gradient-based learning on sequential data, rather than a consequence of specific architectural features. This resilience also explains why transformers with varying configurations achieve comparable language task performance.}

\begin{figure}
    \centering
    \includegraphics[width=.8\linewidth]{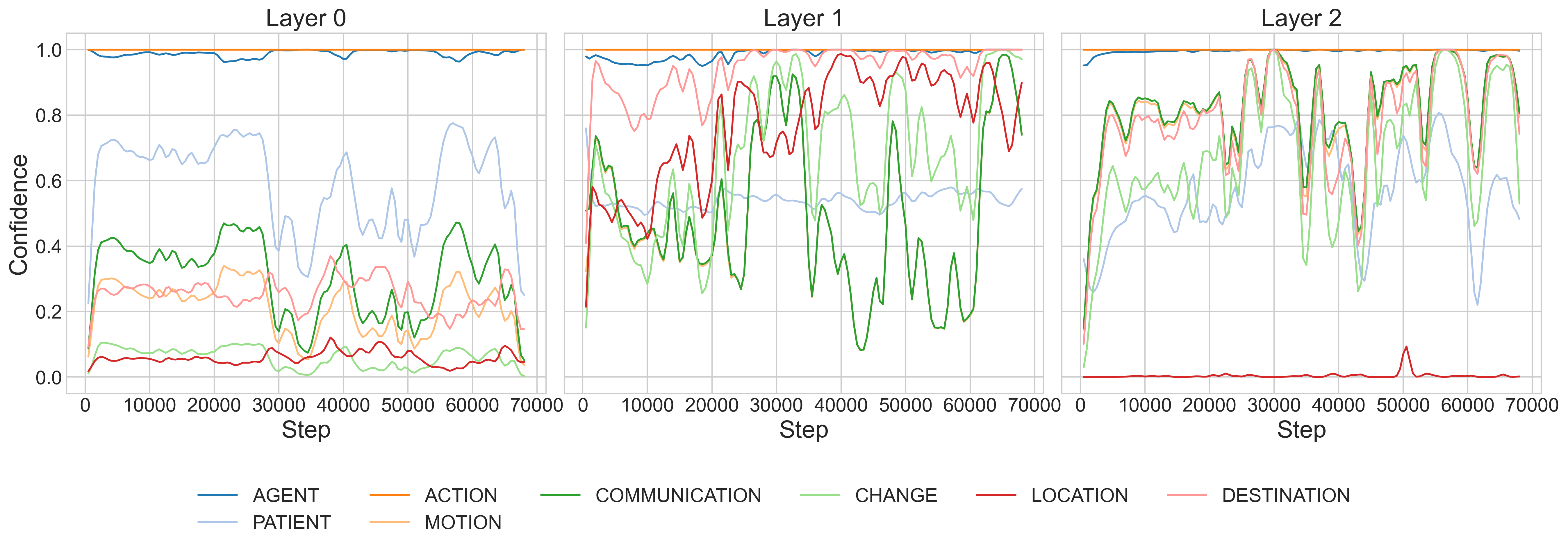}
    \caption{Probe confidence scores over training steps for the large model. Each subplot corresponds to a different decoder layer, with curves representing average model confidence for the presence of specific linguistic tags.}
    \label{fig:probe-pos-comparison}
\end{figure}

\subsection{Linguistic Alignment with Geometric Transitions}
The intersection of curvature and ID trajectories coincides with key transitions in linguistic abstraction emergence. Fig.~\ref{fig:probe-pos-comparison} shows how internal representations evolve across decoder layers, measured by probe confidence scores for linguistic categories (semantics). Layer 1 exhibits an interesting pattern of representational reorganisation, evident by a temporary dip and increased volatility in confidence scores, occurring precisely at the curvature-ID intersection shown in Fig.~\ref{fig:curvature-id-comparison}. This suggests global geometric shifts correspond to evolving category structure. 

Layer 2 shows increased category confidence around the same intersection region, mirroring Layer 1's earlier drop, indicating a hand-off dynamic between middle and upper layers. This temporal complementarity reflects upward abstraction shifts, whereby higher layers specialise in more abstract linguistic features. Unlike Layer 0's stable dominance and Layer 1's volatility, Layer 2 maintains a fragmented, dynamic profile throughout training, supporting the formation of higher-order abstractions or interpretability-relevant circuits.

Medium-sized models (Appendix~\ref{app:probebaselines}) show more harmonised behaviour across layers, demonstrating tighter coupling between abstraction capacity and model scale, with similar behaviour to layer 1 in the large model. These results support that evolving probe confidence reflects internal reorganisation aligned temporally with geometric transitions.

\begin{figure}[]
    \centering
    \includegraphics[width=.8\textwidth]{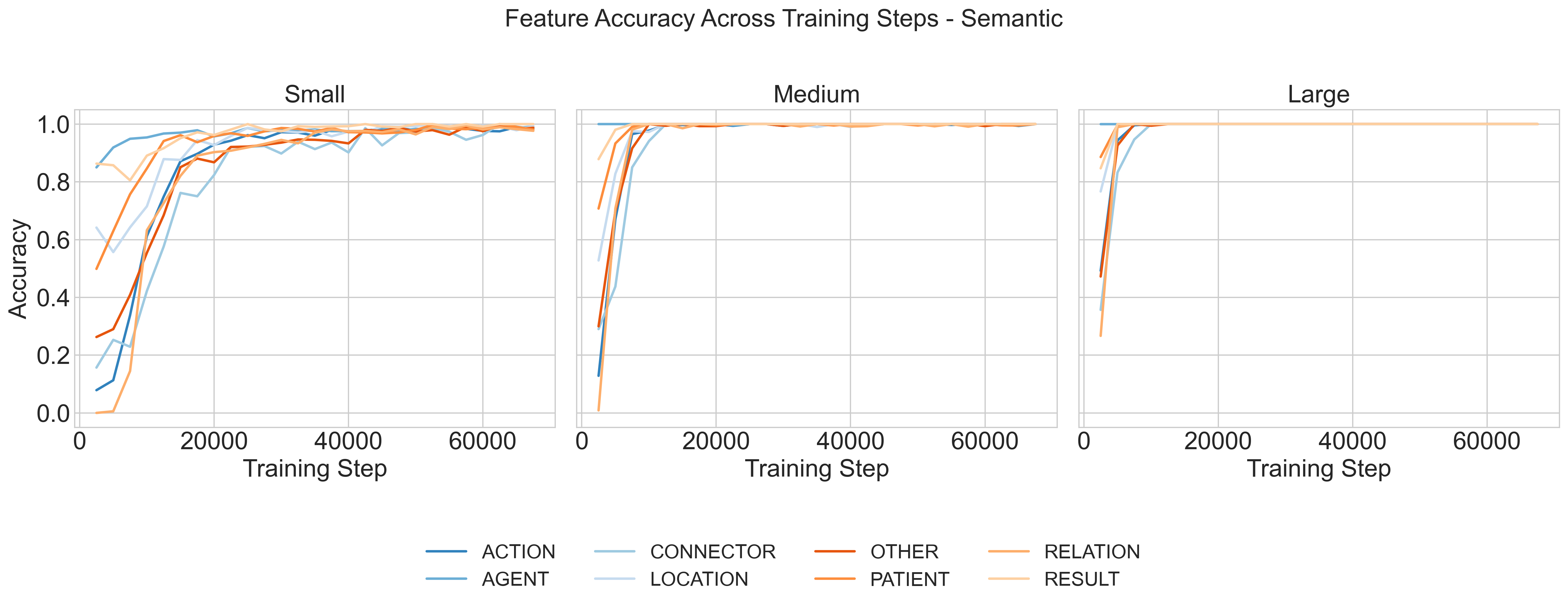}
    \caption{SRL performance per label across models and training steps}
    \label{fig:output-pos-comparison}
\end{figure}

This internal development diverges from output predictions (Fig.~\ref{fig:output-pos-comparison}), where semantic classification accuracy rapidly improves and stabilises early in training. This indicates that while models quickly generate syntactically appropriate tokens, internal representations continue restructuring long after. This dissociation implies a two-phase developmental process: in the first phase, output behaviour reflects coarse category distinctions likely driven by surface-level statistical regularities; in the second, deeper abstraction is gradually encoded into the model's internal geometry, as indicated by evolving probe confidence. Full results across models (Appendix~\ref{app:probes_labels}) show consistent probe dynamics for both syntactic and semantic categories.

\subsection{Limitations of Statistical Diagnostics}
Despite theoretical appeal, MI analysis failed to yield reliable insights in our experimental setting. The observed MI dynamics exhibited high variance and showed minimal alignment with phase transitions identified through geometric and linguistic metrics. This instability aligns with concerns raised by \citet{aljaafari2025carma}, which argues that such patterns in transformers may reflect stochastic fluctuations in early representation formation rather than meaningful abstraction signals. Given these limitations, we do not report MI in the main body of this paper, though complete MI trajectories are reported in Appendix \ref{app:mine}.
\section{Discussion and Conclusion}
Our results suggest that transformers undergo structured representational reorganisation during training. Rather than emerging gradually, abstraction is evident through phase transitions, coordinated shifts across geometric, information-theoretic, and linguistic signals. These transitions mark a distinct boundary between memorisation and generalisation, where linguistic representations begin to stabilise.

We observe consistent alignment between curvature flattening, ID rise and stabilisation, and increased probe accuracy. This coordination suggests certain geometric signals may serve as markers of emerging abstraction. Low-rank curvature and stable ID appear to signal when models begin internalising structure beyond surface patterns. The phenomenon remains robust across model scales and architectural variants, suggesting that abstraction emergence follows predictable patterns. 

While our experiments use synthetic corpora, TRACE is compatible with broader domains. Metrics such as curvature and ID are model-agnostic and can be applied to pre-trained transformers or fine-tuning regimes. Probing-based signals can be approximated using weak supervision or automated annotation tools. Extending TRACE to large-scale pretraining could reveal whether similar phase transitions emerge in noisier, real-world settings. Finally, integrating TRACE with mechanistic interpretability tools could help localise where and how abstraction-related circuits emerge.
\section{Limitations}\label{sec:limitations}
Despite its interpretability, our synthetic corpus does not fully capture the ambiguity and richness of NL.  While probe-based diagnostics offer valuable insights, they provide a static view of representation content and may not reflect the dynamic computational mechanisms that transformers deploy at inference time. Finally, while TRACE establishes strong correlations across geometric, informational, and linguistic signals, it does not establish causal relationships or quantify the relative contribution of each factor.

\section{Impact Statement}\label{sec:impact_statement}
This work reveals that abstract reasoning in language models emerges through predictable phase transitions rather than gradual accumulation. Identifying these critical transitions could enable more efficient training strategies, yielding more interpretable models with reduced computational costs. While this understanding may help detect harmful behaviours, it also presents the usual interpretability trade-off of potentially facilitating model manipulation. Our frame-semantic data generation framework provides a reusable tool for studying abstraction and learning dynamics in language models, with fine-grained control over linguistic properties and transparent evaluation capabilities.

\bibliographystyle{plainnat}
\bibliography{references}

\newpage

\appendix
\section{Frame-Semantic Data Generation Framework}\label{app:synthetic-data}
This appendix describes our controllable synthetic data generation framework \textbf{ABSynth}. Unlike template-based synthetic datasets~\citep{lake2018generalization} or task-specific benchmarks~\citep{saxton2019analysing}, ABSynth is grounded in formal frame semantics~\citep{fillmore1982frame, baker-etal-1998-berkeley-framenet} and is able to generate English-like corpora by \textit{sampling} from abstract event frames with predefined semantic roles, \textit{enabling mechanistic study of how linguistic abstractions emerge during transformer training.}

\begin{figure}[h]
    \centering
    \includegraphics[width=\linewidth]{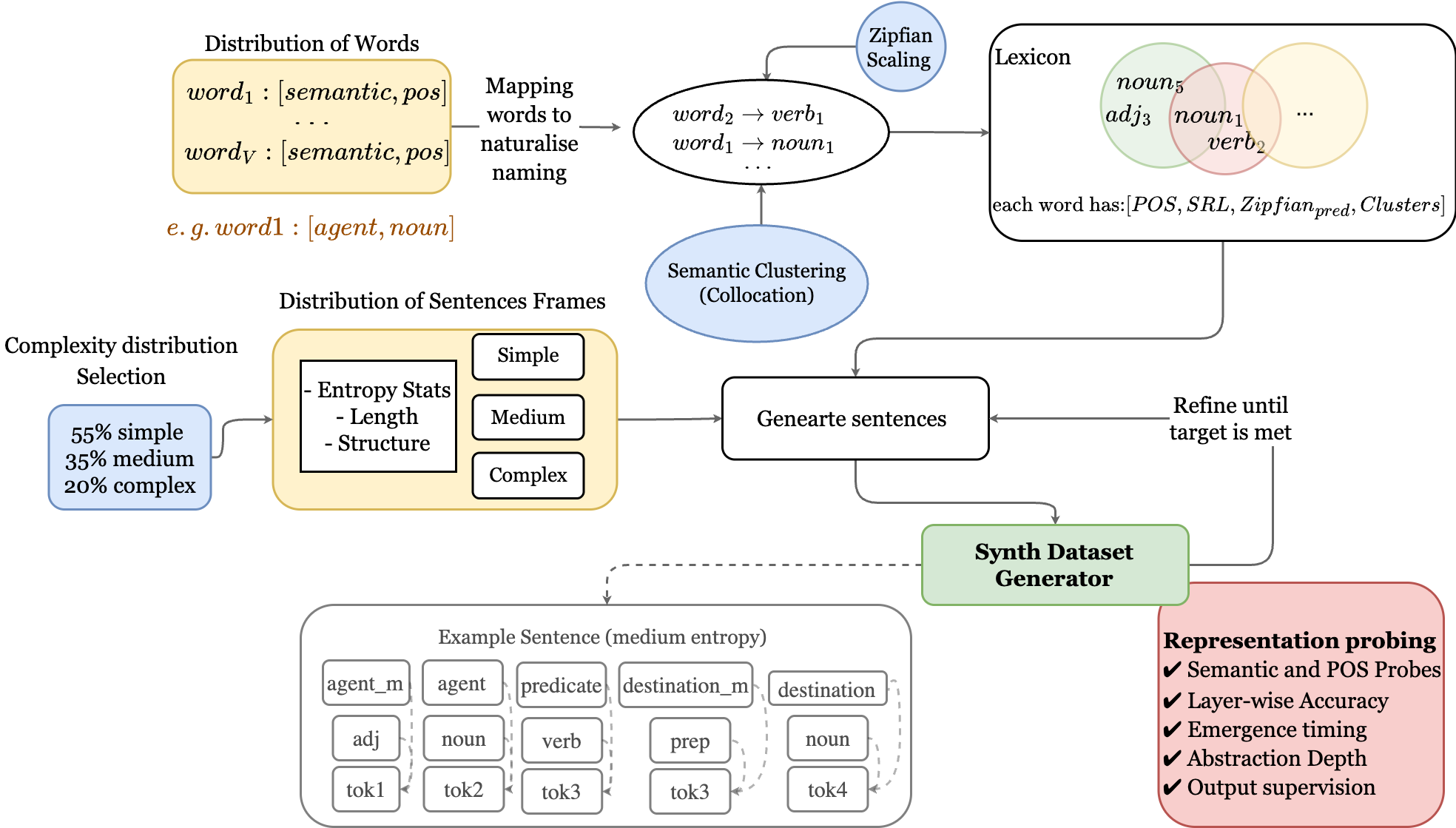}
    \caption{The frame-semantic data generation pipeline: (1) Frame selection with semantic roles, (2) Lexical realisation with Zipfian scaling, (3) Syntactic construction following grammatical constraints, (4) Entropy calibration for controlled predictability. Each generated sentence preserves ground-truth annotations from the underlying frame structure.}
    \label{fig:absynth_diagram}
\end{figure}

ABSynth operationalises frame semantics by representing events as structured predicate-argument frames. Each frame specifies: (i) frame elements; (ii) core semantic roles (e.g., \texttt{AGENT}, \texttt{PATIENT}, \texttt{INSTRUMENT}); (iii) corresponding syntactic categories (e.g., \texttt{NOUN}, \texttt{VERB}), and (iv) complexity constraints based on sequence length and entropy calibration.

This grounding ensures that generated sentences exhibit genuine compositional structure rather than arbitrary token sequences. The frame-based approach enables direct tracking of how neural models learn to represent abstract semantic categories that underlie surface forms.


As illustrated in Figure~\ref{fig:absynth_diagram}, ABSynth generates datasets through a multi-stage pipeline that includes: (i) semantic frame selection with role specification, (ii) lexical realization following Zipfian frequency distributions and semantic clustering, (iii) syntactic construction respecting grammatical constraints and frame-to-syntax mappings, and (iv) entropy calibration to control contextual predictability. The resulting corpora exhibit theoretically-grounded compositional structure while maintaining naturalistic statistical properties.

We detail the generation process below using our ABSynth25K instantiation as an example, emphasising that each component is modular and configurable for different research objectives. This flexibility enables systematic exploration of how specific linguistic properties influence abstraction emergence in neural models.

\subsection{Lexicon Construction: Scaling and Semantic Clustering}

The lexicon is built by assigning tokens to POS and semantic role categories, then applying Zipfian scaling and semantic clustering. Each token is associated with a tuple \texttt{[POS, SRL, Zipf\_rank, ClusterID]}, ensuring interpretability and alignment across analysis stages.

To mimic natural lexical statistics, token frequencies follow a Zipfian distribution \citep{zipf1949human, piantadosi2014zipf}:
\begin{equation}
P(tok_i) = \frac{1/i^{\alpha} + \varepsilon_i}{\sum_{j=1}^{V} (1/j^{\alpha} + \varepsilon_j)},
\end{equation}
where $\alpha = 1.05$, $\varepsilon_i \sim \mathcal{N}(0, 0.05)$, and $V = 9000$ is the vocabulary size. This maintains a realistic long-tail frequency spectrum.

Semantic clustering introduces collocational structure by forming token groups with variable intra- and inter-cluster association strengths. For tokens $tok_i$ and $tok_j$ belonging to clusters $c_i$ and $c_j$:
\begin{equation}
S(tok_i, tok_j) = S_{base}(c_i, c_j) + \mathcal{U}(0, S_{range}(c_i, c_j)),
\end{equation}
where $S_{base}$ and $S_{range}$ are tuned to create structured but noisy associations, emulating semantic co-occurrence patterns. Intra-cluster associations are drawn from $[0.4, 0.7]$, while cross-cluster links are weaker $([0.05, 0.2])$.

Each word receives a naturalised name (e.g., \texttt{result1}, \texttt{noun5}), allowing transparent reverse-mapping for analysis.

\subsection{Frame-Based Syntactic Realisation and Entropy Control}

Frame-to-syntax mappings define how semantic frames are realised as surface forms while controlling contextual predictability through entropy calibration. All realisations follow valid English grammatical constructions, respecting part-of-speech ordering, agreement patterns, and canonical phrase structure. This grounding enables the study of syntactic and semantic abstraction within structurally coherent input sequences.

Each frame component is annotated with expected entropy based on its role in the frame: 

\begin{itemize} 

    \item \textbf{Low entropy} (0.5–1.5 bits): Grammatically determined positions (e.g., determiners required by nouns) 
    \item \textbf{Medium entropy} (1.5–3.0 bits): Semantically constrained positions with multiple valid fillers (e.g., theme roles that accept various object types) 
    \item \textbf{High entropy} (3.0–4.5 bits): Optional frame elements with high variability (e.g., adverbial modifier) 

\end{itemize}

Frames are instantiated according to a target complexity distribution (55\% simple, 35\% medium, 10\% complex), which guides the global entropy profile of the corpus. Example frame realisations include:

\begin{verbatim}
Simple TRANSFER:    [AGENT=NOUN] [ACTION=VERB] [THEME=NOUN]
                   "noun2 verb3 noun5"

Medium CREATION:    [AGENT=NOUN] [ACTION=VERB] [THEME=NOUN] 
                   [PURPOSE=PREP+NOUN]
                   "noun1 verb2 noun4 prep3 noun7"

Complex MOTION:     [AGENT=NOUN] [REL] [ACTION=VERB] [SOURCE=NOUN]
                   [ACTION=VERB] [GOAL=NOUN]
                   "noun3 rel1 verb5 noun6 verb7 noun9"
\end{verbatim}

\subsection{Dynamic Entropy Adjustment Algorithm}

To enforce statistical balance across complexity levels, the sentence generator incorporates an entropy-aware sampling mechanism. During generation, the system maintains a global entropy profile, defined as the frequency distribution of sentence positions assigned to low, medium, and high entropy tiers. This profile is updated in real time and compared to the desired target distribution.

If the observed distribution diverges from the target (e.g., too many low-entropy tokens have been sampled), the system increases the sampling weight for frames or token categories that contribute to underrepresented tiers. This feedback mechanism modulates the difficulty of the dataset without sacrificing grammaticality.

\begin{algorithm}
\caption{Entropy-Calibrated Sentence Generation}\label{algorithm:synthitc_data}
\begin{algorithmic}[1]
\FOR{each sentence to be generated}
    \STATE Select template $T$ with entropy tier annotations
    \FOR{each token slot $i$ in $T$}
        \STATE Compute current global entropy profile $E_{current}$
        \STATE Compare to target profile $E_{target}$
        \STATE Derive correction factor $\alpha$ to adjust token sampling
        \STATE Sample token $tok_i$ from adjusted distribution
    \ENDFOR
    \STATE Add sentence to corpus
    \STATE Update $E_{current}$ with entropy tags from new sentence
\ENDFOR
\end{algorithmic}
\end{algorithm}

The final vocabulary consists of 9,000 tokens distributed across categories as shown in Table~\ref{tab:voca_dist}.

\subsection{Output Format and Probing Supervision}

Each sentence is stored with a naturalised token sequence and associated structured annotations, including:
\begin{itemize}
    \item \textbf{POS labels} (e.g., \texttt{NOUN}, \texttt{ADJ}, \texttt{VERB})
    \item \textbf{Semantic roles} (e.g., \texttt{AGENT}, \texttt{PATIENT}, \texttt{RESULT})
    \item \textbf{Entropy tier} 
    \item \textbf{Other Contextual complexity metadata}
\end{itemize}

These annotations enable direct supervision for probing tasks. During model training, hidden states are extracted layer-wise and evaluated using linear probes trained on these annotations. This setup facilitates fine-grained analysis of how and when compositional representations emerge, and how they correlate with curvature, intrinsic dimensionality, and mutual information.
\begin{table}[h]
\centering
\begin{tabular}{lr}
\hline
\small
\textbf{Token Category} & \textbf{Vocabulary Size} \\
\hline
Noun & 2,780 \\
Transitive Verb & 694 \\
Intransitive Verb & 694 \\
Communication Verb & 347 \\
Motion Verb & 347 \\
Adjective & 1,388 \\
Adverb & 555 \\
Location & 694 \\
Temporal & 694 \\
Preposition & 416 \\
Determiner & 111 \\
Conjunction & 277 \\
Result & 277 \\
\hline
\textbf{Total} & \textbf{9,000} \\
\hline
\end{tabular}
\caption{Vocabulary distribution across syntactic categories}\label{tab:voca_dist}
\end{table}

\section{Technical Implementation Details}\label{app:implemenation_details}
\subsection{Main Model Architecture}\label{app:model-arch}
We implemented decoder-only transformer architectures based on the original design of \citet{vaswani2017attention}. Each model consists of $L$ layers, each with hidden size $d_{\text{model}}$, $H$ attention heads (where $d_{\text{head}} = d_{\text{model}} / H$), and a feedforward dimension $d_{\text{ffn}}$. We focus on decoder-only models given their increasing prevalence in production LLMs and recent arguments that causal architectures provide a cleaner demonstration for emergence~\citep{skean2025layer}.

Our architectural choices are guided by recent Transformer scaling laws, notably those articulated by \citet{hoffmann2022empirical}. Namely, we evaluate three configurations that adhere to the Chinchilla scaling law (see below), with approximate parameter counts and architectural specifications as presented in Table \ref{tab:model-configs}. All models are trained with a maximum sequence length of $T = 16$ tokens and a dropout rate of $0.1$. 

We use a simple whitespace-based tokeniser, where each token corresponds to a space-separated word or symbol. This choice allows us to maintain interpretability and simplify downstream representational analyses, while aligning with our controlled, low-scale experimental setup.


\begin{table}[H]
  \caption{Transformer model configurations used in this study}
  \label{tab:model-configs}
  \centering
  \begin{tabular}{lcccc}
    \toprule
    \textbf{Model} & \textbf{Layers ($L$)} & \textbf{Hidden size ($d_{\text{model}}$)} & \textbf{Heads ($H$)} & \textbf{FFN size ($d_{\text{ffn}}$)} \\
    \midrule
    Small  & 1 & 64  & 2 & 128 \\
    Medium & 2 & 96  & 3 & 384 \\
    Large  & 3 & 128 & 4 & 512 \\
    \bottomrule
  \end{tabular}
\end{table}

\paragraph{Scaling Laws and Training Budget.}
We follow the Chinchilla scaling principles from \citet{hoffmann2022empirical}, which demonstrate that model size and training tokens should scale together. Specifically, Chinchilla findings show that optimal training requires approximately 20 tokens per parameter. Given our dataset size of approximately 360K tokens per epoch, we designed our training regimen to respect these scaling principles. Based on our model sizes (110K, 339K, and 749K parameters), we estimated minimum training requirements of 6, 19, and 42 epochs, respectively, to ensure a sufficient token-to-parameter ratio, as implied by Chinchilla's 20:1 guideline. However, in practice, we observed that phase transitions occurred at different points across scales, not precisely aligned with these theoretical estimates. As such, we extended training durations beyond these minimums to allow sufficient time for representational transitions to emerge, as discussed in Section \ref{app:training}, and to test whether the Grokking phoneme \citep{power2022grokking} existence.

\subsection{Main Models Training Objectives and Formalisation of Next Token Prediction} 

\paragraph{Next-Token Prediction Task (NTP):}  We formalise the \textit{NTP} task, which we use to train our decoder-only language models. NTP involves predicting the next token \( x_{n+1} \) given a preceding sequence \( X = \{x_1, x_2, \dots, x_n\} \). In our setting, each \( x_i \in \mathcal{V} \) belongs to a controlled lexicon from \textsc{ABSynth}, reflecting structured linguistic categories (e.g., \texttt{noun3}, \texttt{verb2}, etc.) and conforming to correct English grammar.

Formally, the objective is:
\begin{equation}
x_{n+1} = \arg\max_{x \in \mathcal{V}} P(x \mid x_1, x_2, \dots, x_n)
\end{equation}

Here, \( \mathcal{V} \) denotes the model’s token vocabulary, which includes synthetic labels representing part-of-speech categories and their variants (e.g., \texttt{noun1}, \texttt{adj2}). The model autoregressively generates one token at a time, conditioned solely on the preceding tokens.

\textbf{Example Prompt:}  
Given the sequence: \texttt{"noun1 verb2 adj1"}, the task requires predicting the next token, such as: \texttt{"noun3"}, depending on the dataset’s underlying syntactic or semantic generation rules.

We focus on NTP because it reflects the core objective used in many widely adopted pretrained language models (e.g., GPT-style models), making it a natural and effective setting for examining representational and interpretive behaviours in a controlled environment.

\subsection{Main Models Training Configuration}\label{app:training}
All models are trained using next-token prediction with the Adam optimiser \citep{kingma2014adam} with learning rate $1e-3$, and 1000 warm-up steps. Training was conducted for significantly extended epochs beyond the computed minimum requirements of the discussed scaling laws (Section~\ref{app:model-arch}), consistent with the observations of \citet{nanda2023progress} on the "grokking" phenomenon, where generalisation emerges abruptly after an initial memorisation phase. We consider this an extension of the training strategies outlined in \citet{csordas2021devil}, which emphasise the importance of avoiding early stopping to fully exploit the learning capacity of neural models.

Specifically, we trained:
\begin{itemize}
  \item \textbf{Small model}: 500 epochs ($\sim$82× the minimum of 6 epochs)
  \item \textbf{Medium model}: 500 epochs ($\sim$26× the minimum of 19 epochs)
  \item \textbf{Large model}: 500 epochs ($\sim$12× the minimum of 42 epochs)
\end{itemize}

We record dense checkpoints throughout training (every 500 training steps), extracting hidden states and gradients from all layers to compute our diagnosis metrics.

We applied minimal regularisation and continuously monitored \textit{loss, accuracy, gradient dynamics, representational similarity, and mutual information (MI)} throughout training to detect potential phase transitions. This methodological rigour enables us to assess the relationship between model structure and task complexity under controlled and interpretable conditions, while ensuring that all models are trained in accordance with modern scaling principles. To ensure correctness and reproducibility, all experiments were repeated at least 5 times with different random seeds, and we reported the averaged results. 

\subsection{Model performance}
\subsection{Model Performance}
All models demonstrate strong performance on the downstream generation task after their respective phase transitions. Table~\ref{tab:model_performance} summarises the evaluation metrics across model scales.

\begin{table}[h]
\centering
\caption{Performance metrics for small, medium, and large models after phase transition.}
\label{tab:model_performance}
\begin{tabular}{lcccc}
\toprule
\textbf{Model} & \textbf{Exact Match} & \textbf{Token Accuracy} & \textbf{BLEU Score} & \textbf{Perplexity} \\
\midrule
Small & 0.84 & 0.98 & 0.20 & 1.22 \\
Medium & 0.98 & 0.99 & 0.21 & 1.08 \\
Large & 0.98 & 0.99 & 0.21 & 1.07 \\
\bottomrule
\end{tabular}
\end{table}

\subsection{Probing Framework and Label Construction}\label{app:probes_labels}
To investigate the interpretability and internal structure of our models, we implemented a probing framework that trains lightweight classifiers on the frozen hidden representations extracted from each layer of our trained models. Specifically, for every layer in the tested models, we trained and evaluated both a part-of-speech (POS) probe and a semantic role labelling (SRL) probe.

Each probe is implemented as a feedforward neural network comprising three linear layers interleaved with ReLU activations and dropout. The network is trained using binary cross-entropy loss, with sigmoid activations at the output layer to support multi-label classification. Given an input representation $\mathbf{h} \in \mathbb{R}^d$, the probe computes:

\[
\hat{\mathbf{y}} = \sigma\left(W_3 \cdot \text{ReLU}\left(W_2 \cdot \text{ReLU}\left(W_1 \cdot \mathbf{h}\right)\right)\right),
\]
where $W_1, W_2, W_3$ are trainable weight matrices and $\sigma$ denotes the element-wise sigmoid function.
\begin{table}[H]
\centering
\small
\caption{POS Probe Evaluation – Large Model, Layer 0}
\label{tab:pos_probe_layer_0}
\begin{tabular}{lccccc}
\toprule
\textbf{Label} & \textbf{Count} & \textbf{Accuracy} & \textbf{Precision} & \textbf{Recall} & \textbf{F1 Score} \\
\midrule
NOUN & 279984 & 1.000 & 1.000 & 1.000 & 1.000 \\
TRANSITIVE\_VERB & 151232 & 0.836 & 0.577 & 0.836 & 0.683 \\
INTRANSITIVE\_VERB & 59584 & 0.073 & 1.000 & 0.073 & 0.136 \\
COMMUNICATION\_VERB & 54352 & 0.238 & 1.000 & 0.238 & 0.384 \\
MOTION\_VERB & 68368 & 0.344 & 0.659 & 0.344 & 0.452 \\
CHANGE\_VERB & 14240 & 0.562 & 0.510 & 0.562 & 0.535 \\
ADJ & 59072 & 0.239 & 0.535 & 0.239 & 0.331 \\
LOCATION & 101984 & 0.315 & 0.858 & 0.315 & 0.461 \\
TEMP & 37664 & 0.184 & 0.535 & 0.184 & 0.274 \\
PREP & 115264 & 0.387 & 0.823 & 0.387 & 0.526 \\
RESULT & 68592 & 0.292 & 0.746 & 0.292 & 0.420 \\
CONJ & 51248 & 0.350 & 0.874 & 0.350 & 0.500 \\
\bottomrule
\end{tabular}
\end{table}
Probes are trained independently for each layer in the model, allowing us to analyse the emergence and distribution of linguistic and functional features across depth. Representations from each layer are frozen and taken from an already trained model following our training setting explained in Section~\ref{app:implemenation_details}, and the underlying model weights are not updated during probing.

Due to the synthetic nature of our dataset, both POS and semantic labels are deterministically derived from token names. For example, a token such as \texttt{noun3} is assigned the POS label \texttt{NOUN} and may additionally be annotated with semantic roles such as \texttt{AGENT} or \texttt{ENTITY}, depending on the symbolic structure of the task. This approach eliminates annotation ambiguity and ensures consistent supervision across examples.
\begin{table}[H]
\centering
\small
\caption{POS Probe Evaluation – Large Model, Layer 1}
\label{tab:pos_probe_layer_1}
\begin{tabular}{lccccc}
\toprule
\textbf{Label} & \textbf{Count} & \textbf{Accuracy} & \textbf{Precision} & \textbf{Recall} & \textbf{F1 Score} \\
\midrule
NOUN & 279984 & 1.000 & 1.000 & 1.000 & 1.000 \\
TRANSITIVE\_VERB & 151232 & 0.779 & 0.655 & 0.779 & 0.711 \\
INTRANSITIVE\_VERB & 59584 & 0.291 & 1.000 & 0.291 & 0.451 \\
COMMUNICATION\_VERB & 54352 & 0.476 & 1.000 & 0.476 & 0.645 \\
MOTION\_VERB & 68368 & 0.718 & 0.678 & 0.718 & 0.697 \\
CHANGE\_VERB & 14240 & 0.875 & 0.510 & 0.875 & 0.644 \\
ADJ & 59072 & 0.603 & 0.558 & 0.603 & 0.580 \\
LOCATION & 101984 & 0.641 & 0.843 & 0.641 & 0.729 \\
TEMP & 37664 & 0.344 & 0.535 & 0.344 & 0.419 \\
PREP & 115264 & 0.856 & 0.808 & 0.856 & 0.831 \\
RESULT & 68592 & 0.585 & 0.746 & 0.585 & 0.655 \\
CONJ & 51248 & 0.545 & 1.000 & 0.545 & 0.705 \\
\bottomrule
\end{tabular}
\end{table}
Given the supervised design of our dataset, we employ linear probes due to their demonstrated effectiveness in similar contexts. While alternative methods such as Sparse Autoencoders (SAEs) have shown promise in unsupervised settings \citet{cunningham2023sparse}, linear probes remain a robust and interpretable choice for supervised feature probing \citet{kantamneni2025sparse, smith2025negative}.
\subsection{Probe Models Training Configuration}
All probes were trained for 30 epochs using the Adam optimiser \citep{kingma2014adam} with a learning rate of $1 \times 10^{-4}$, a hidden dimension of 256, and a dropout rate of 0.5. Input dimensionality matched the model's hidden state size.

\subsection{Probing performance}
We report the average POS probe models' performance on Tables \ref{tab:pos_probe_layer_0}, \ref{tab:pos_probe_layer_1}, and \ref{tab:pos_probe_layer_2}, for the small, medium, and large models, receptively. We also report the average semantic probe models' performance on Tables \ref{tab:semantic_probe_layer_0}, \ref{tab:semantic_probe_layer_1}, and \ref{tab:semantic_probe_layer_2}, for the small, medium, and large models, receptively.
\begin{figure}
    \centering
    \includegraphics[width=0.32\linewidth]{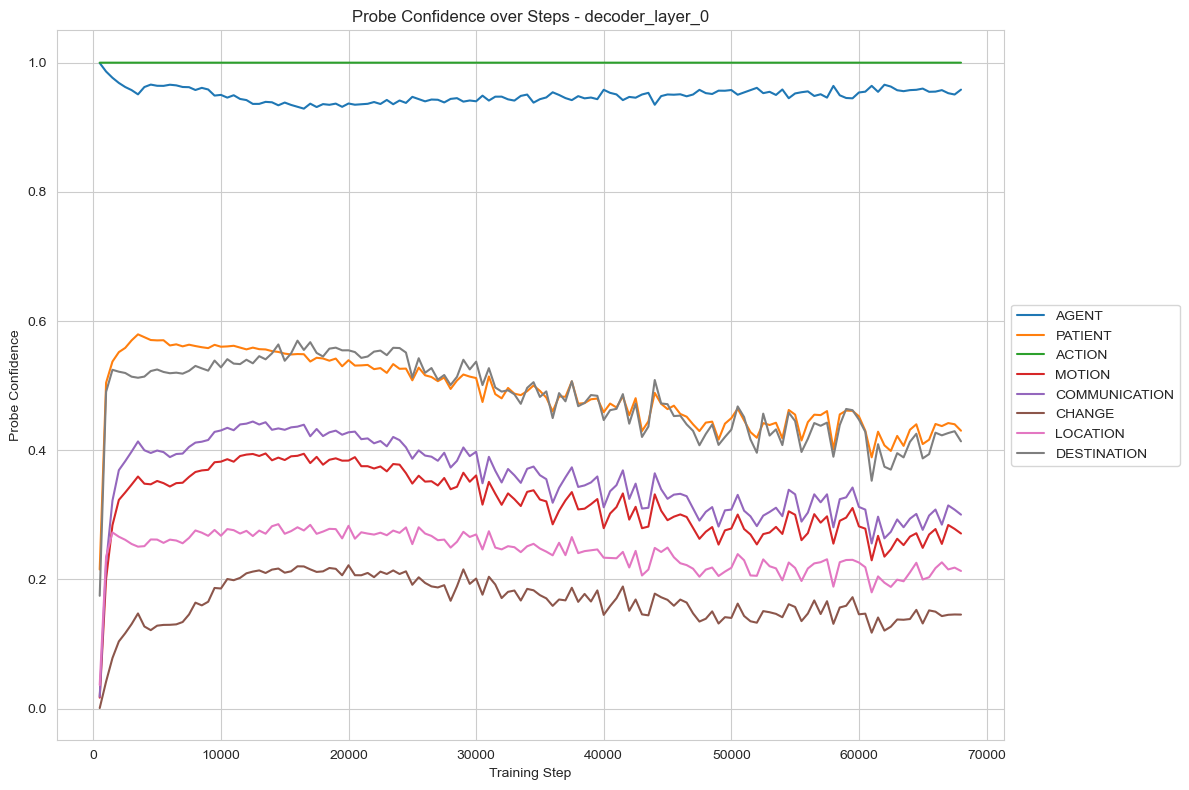}
    \includegraphics[width=0.32\linewidth]{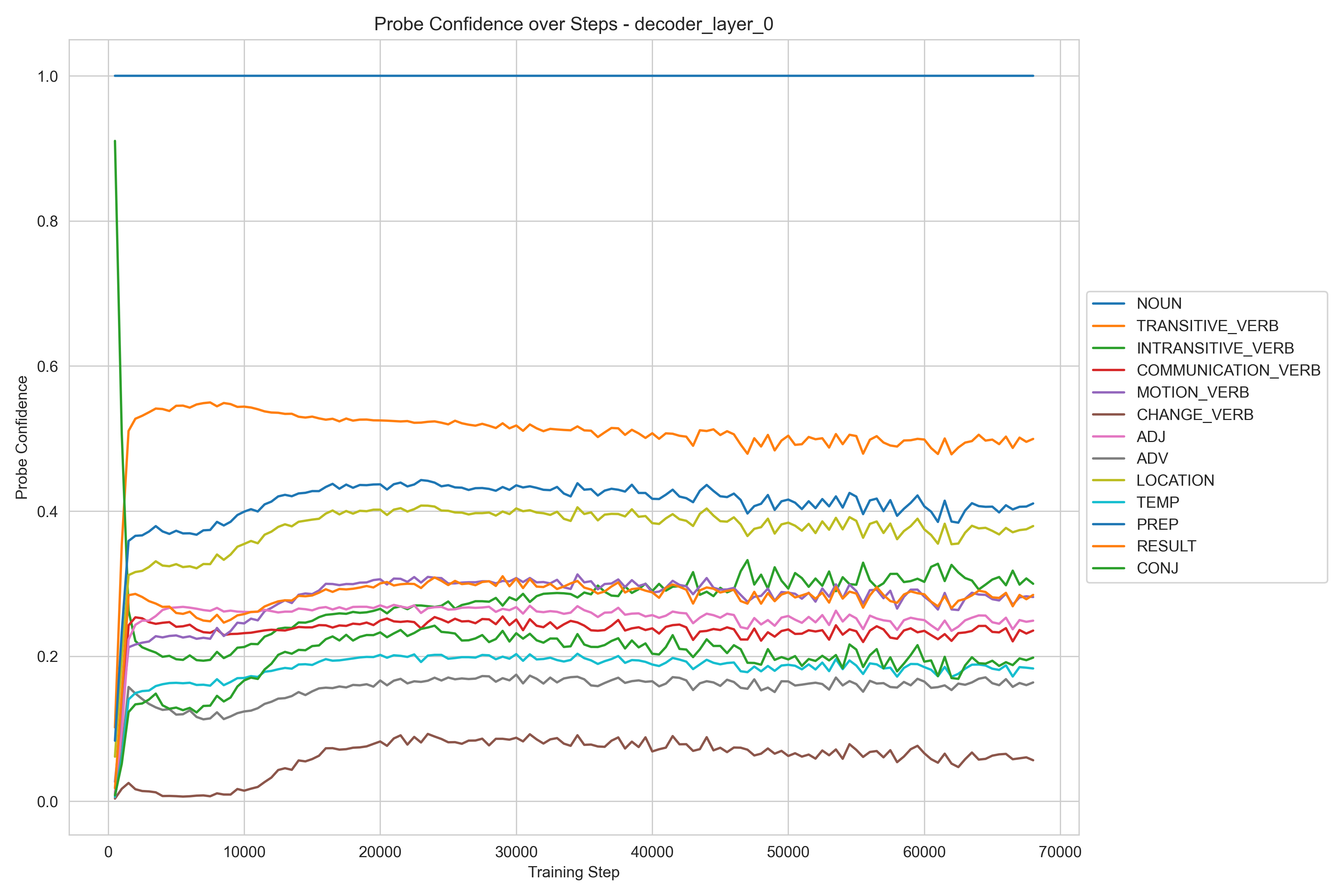}
    \caption{Small model — Semantic (left) and POS (right) probe confidence scores at Layer 0.}
    \label{fig:probe_s}
\end{figure}

\begin{figure}
    \centering
    \includegraphics[width=0.42\linewidth]{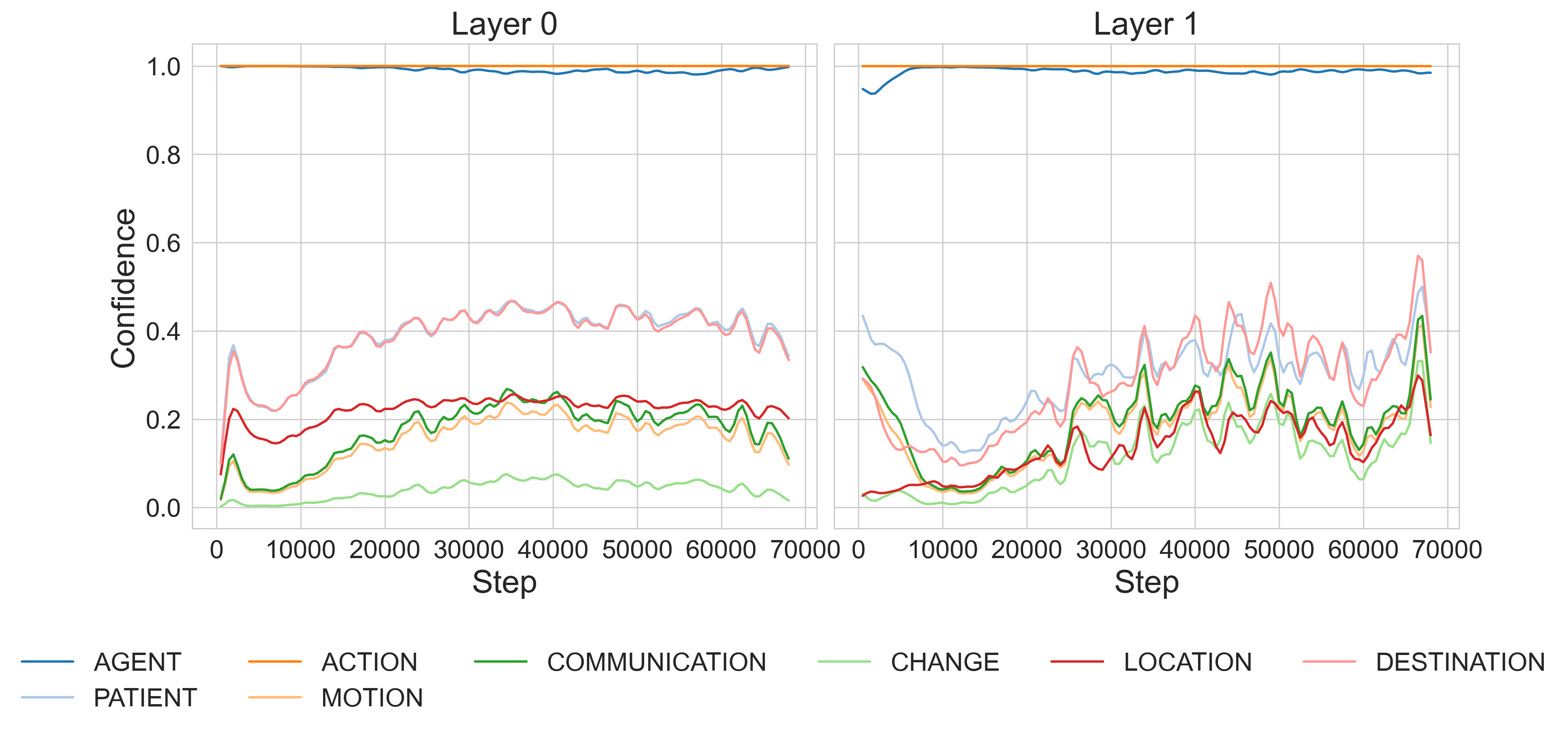}
    \includegraphics[width=0.43\linewidth]{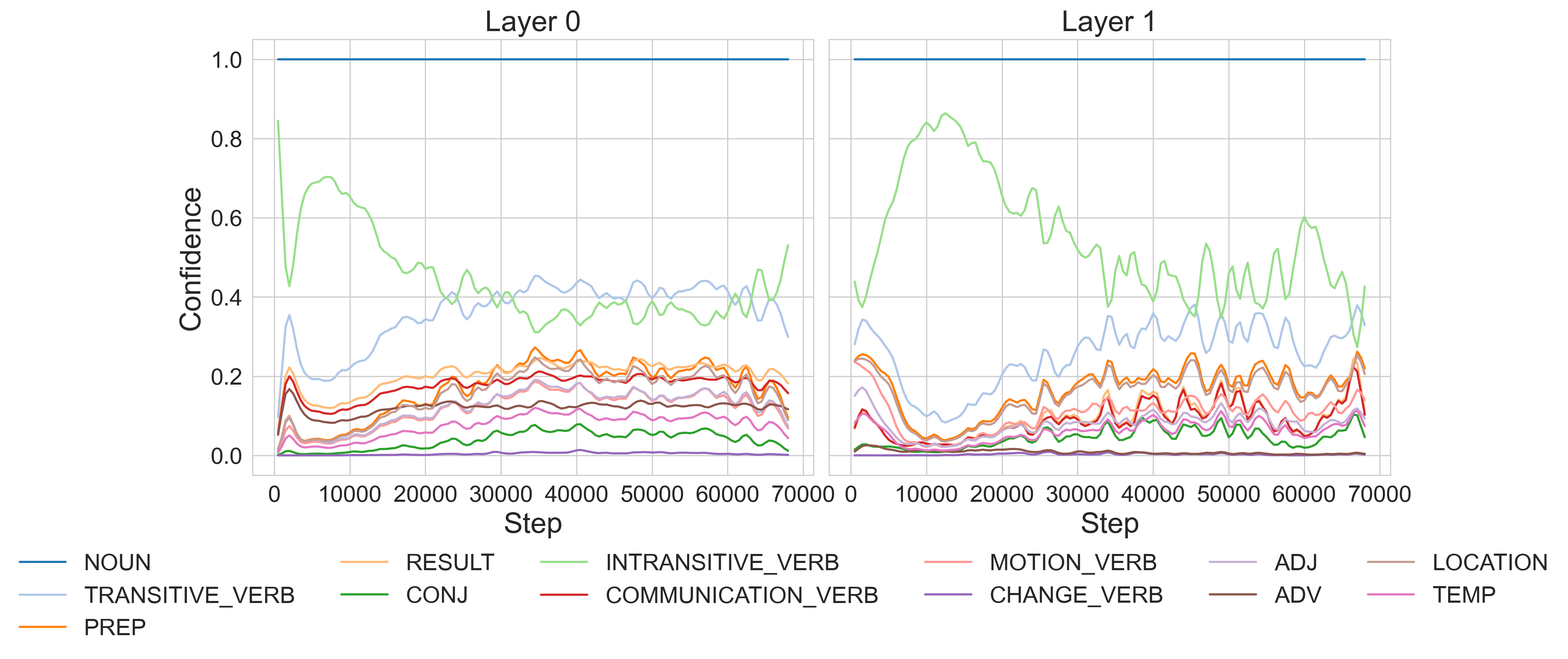}
    \caption{Medium model — Semantic (left) and POS (right) probe confidence scores across layers.}
    \label{fig:probe_m}
\end{figure}

\subsection{Probing Extended Results}\label{app:probebaselines}
\begin{figure}
    \centering
    \includegraphics[width=0.5\linewidth]{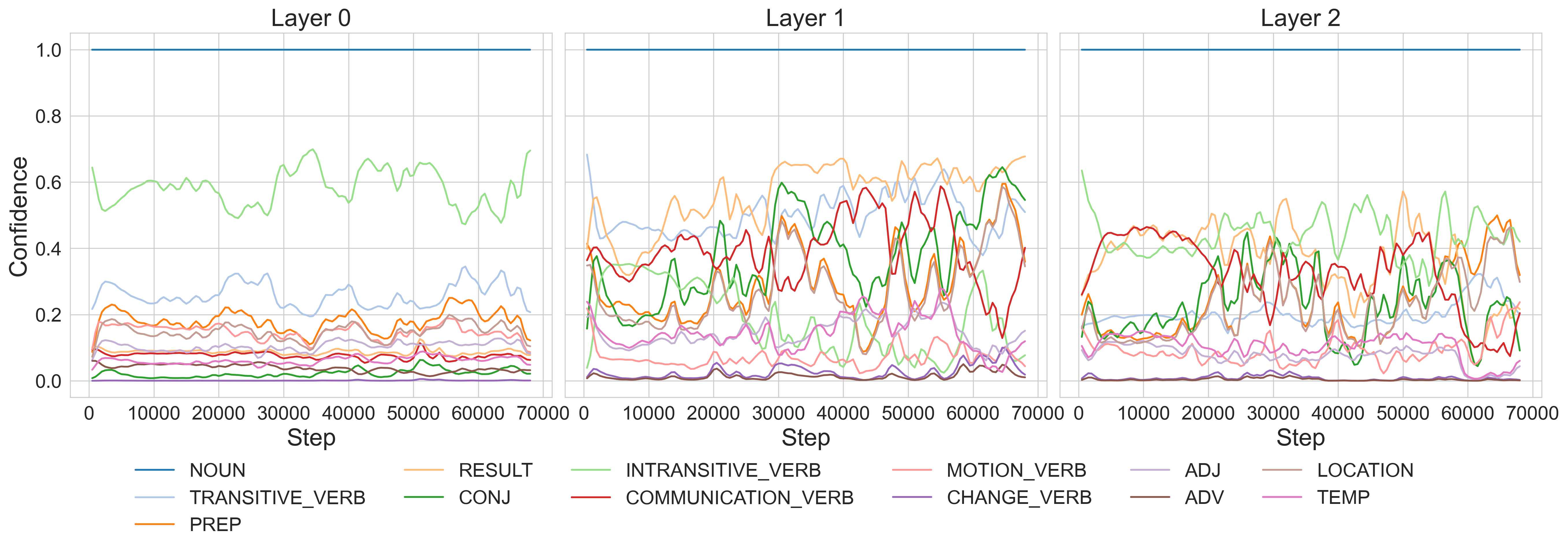}
    \caption{Large model — POS probe confidence scores across layers.}
    \label{fig:probe_l_pos}
\end{figure}

We present extended probing results across all model scales (Small, Medium, and Large) for both part-of-speech (POS) and semantic role categories. Figures~\ref{fig:probe_s},~\ref{fig:probe_m}, and~\ref{fig:probe_l_pos} show model confidence scores for each layer across training steps.


To assess whether abstraction emerges during training rather than being encoded by architecture alone, we also trained probes on frozen hidden states from randomly initialised models. These models exhibited near-zero performance across all linguistic categories, confirming the absence of structured representations at initialisation.

\begin{figure}
    \centering
    \includegraphics[width=0.32\linewidth]{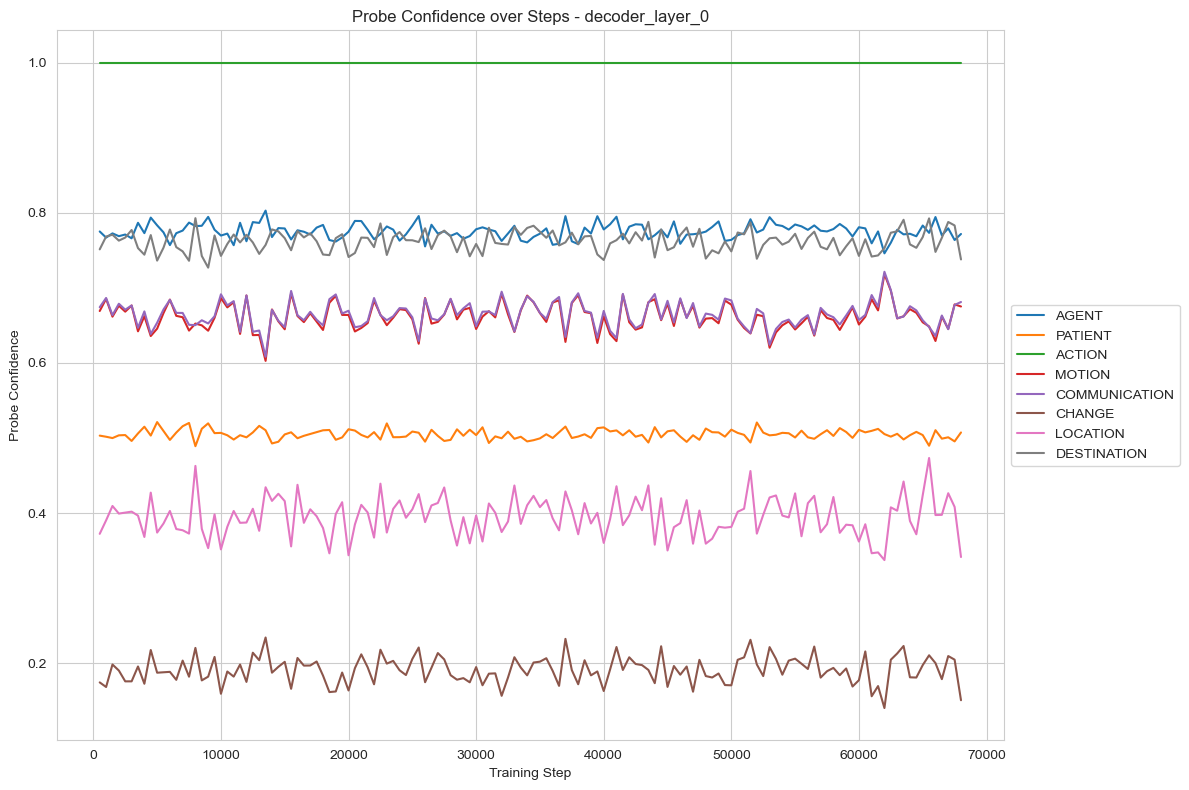}
    \includegraphics[width=0.32\linewidth]{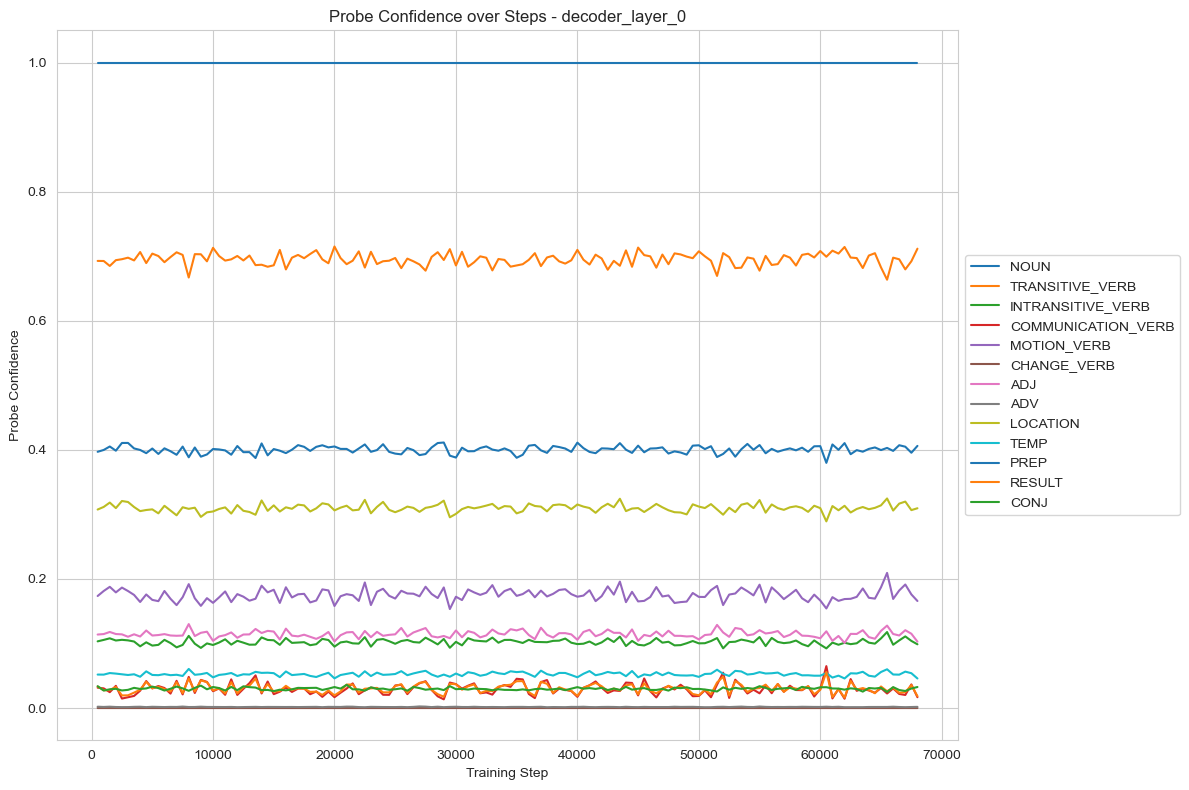}
    \caption{Randomly initialised Small model — Semantic (left) and POS (right) probe confidence scores.}
    \label{fig:probe_s_rand}
\end{figure}

\begin{figure}
    \centering
    \includegraphics[width=0.42\linewidth]{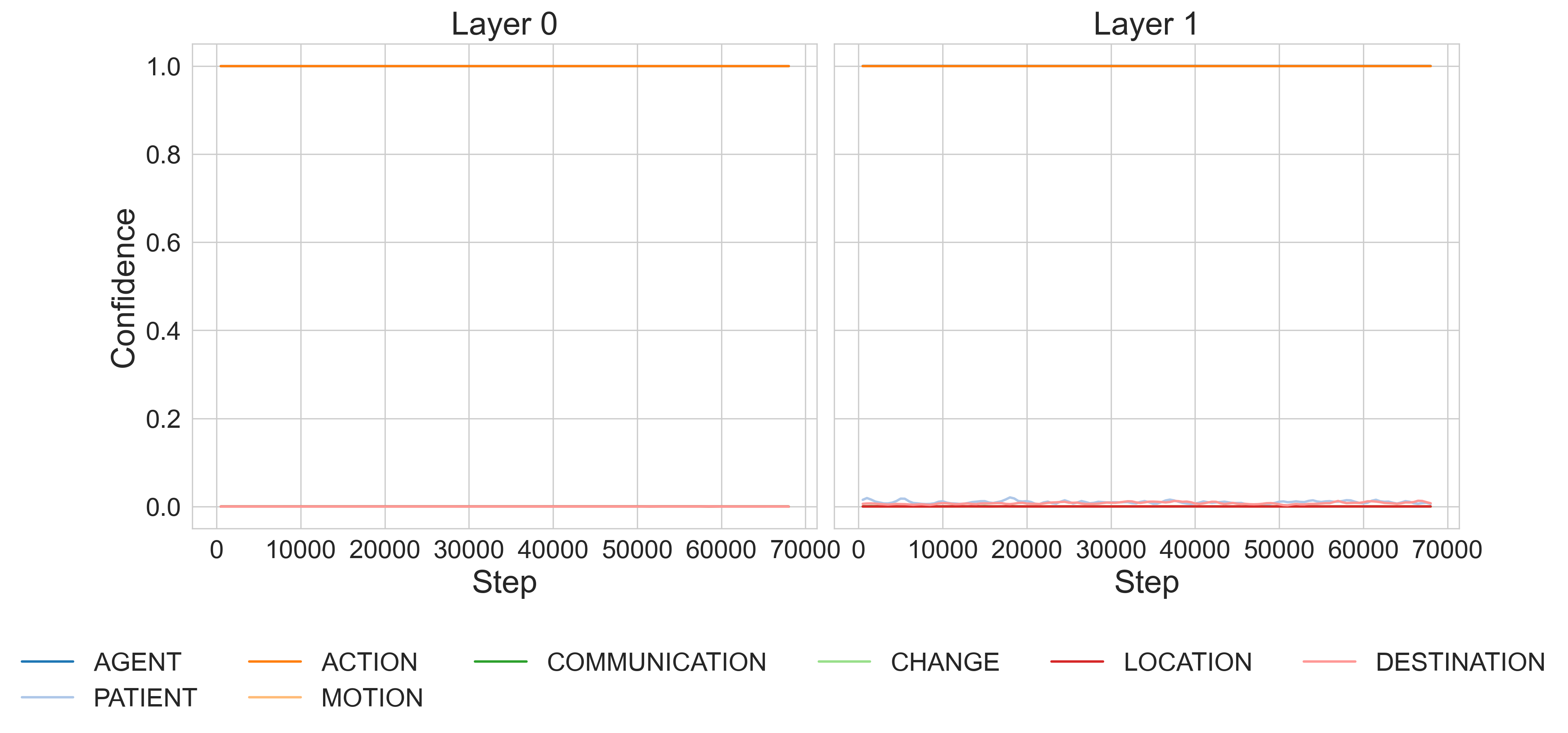}
    \includegraphics[width=0.43\linewidth]{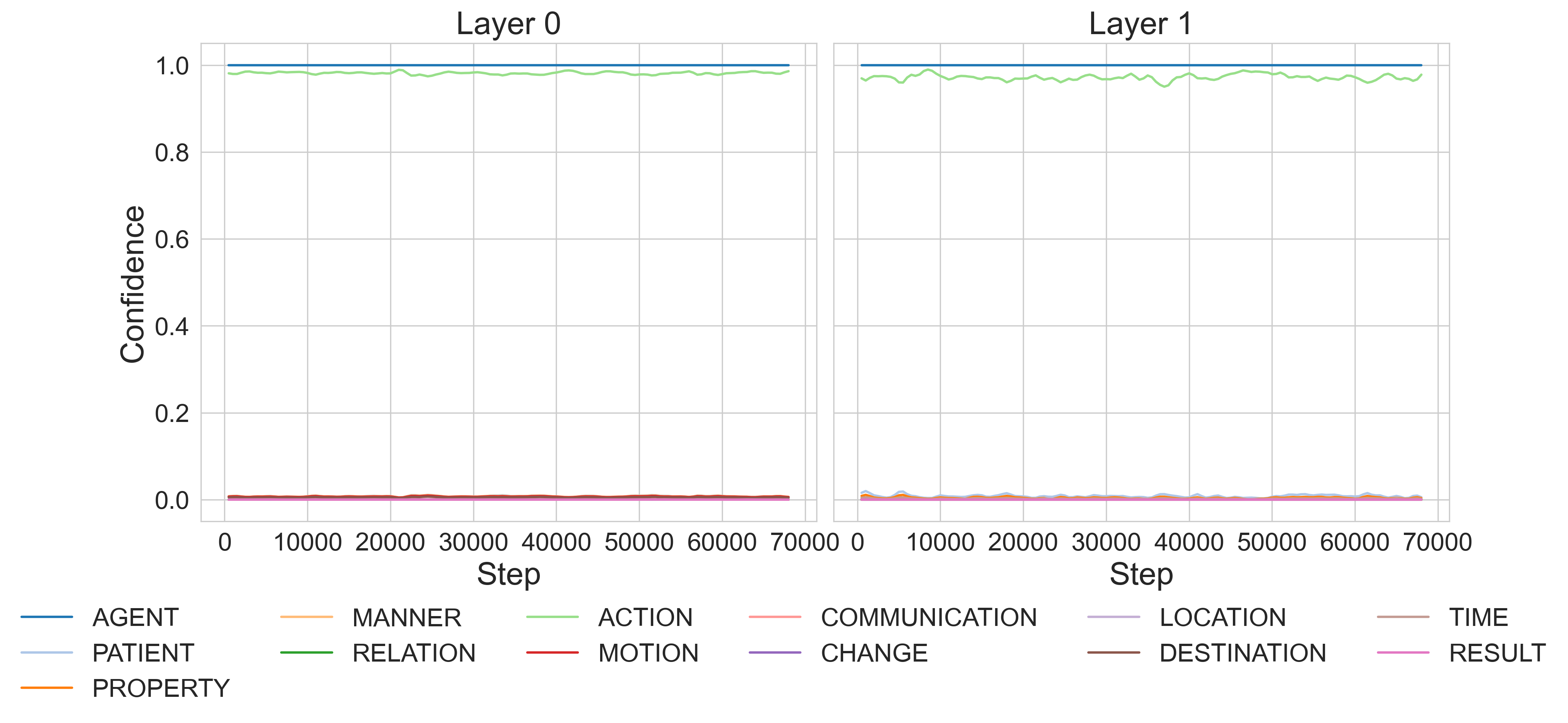}
    \caption{Randomly initialised Medium model — Semantic (left) and POS (right) probe confidence scores.}
    \label{fig:probe_m_rand}
\end{figure}

\begin{figure}
    \centering
    \includegraphics[width=0.42\linewidth]{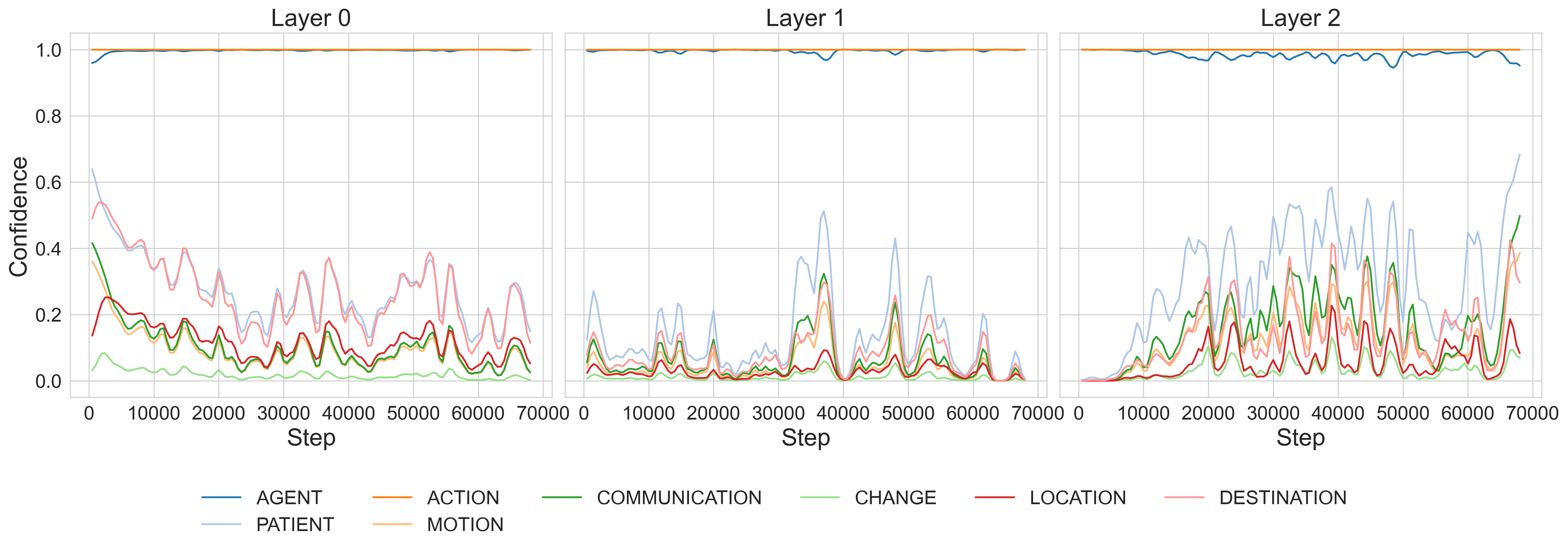}
    \includegraphics[width=0.43\linewidth]{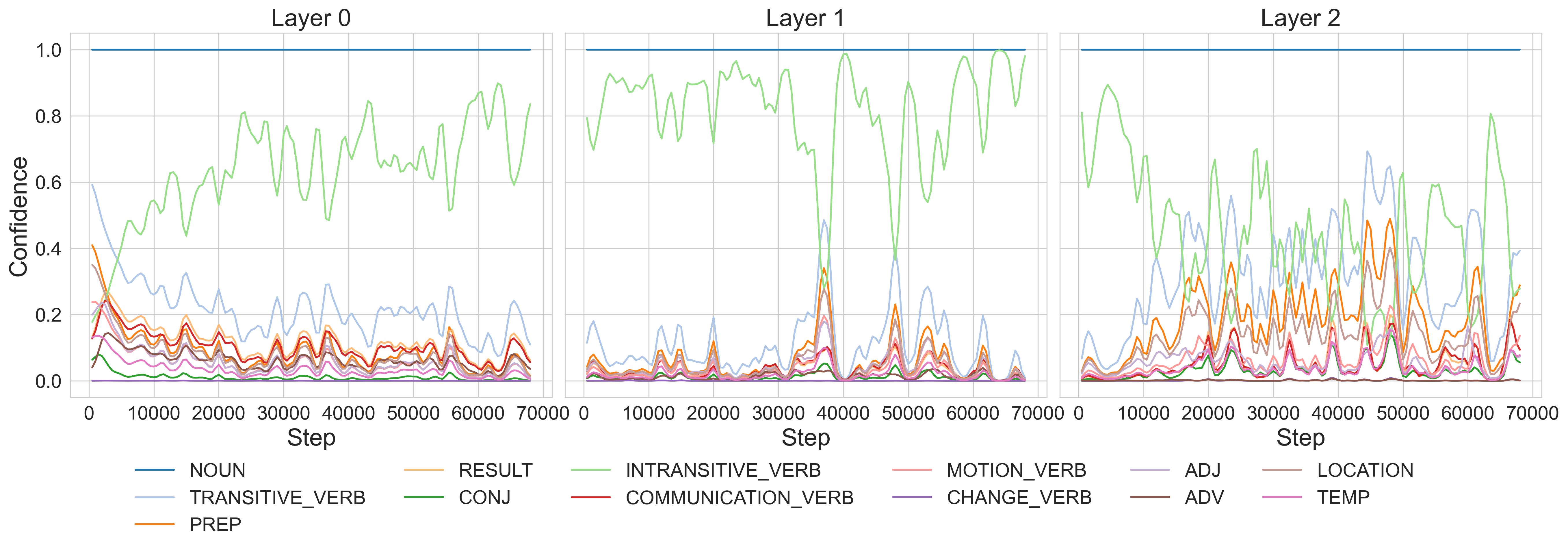}
    \caption{Randomly initialised Large model — Semantic (left) and POS (right) probe confidence scores.}
    \label{fig:probe_l_rand}
\end{figure}

Figures~\ref{fig:probe_s_rand},~\ref{fig:probe_m_rand}, and~\ref{fig:probe_l_rand} show the performance of semantic and POS probes on randomly initialised models.

\begin{table}[H]
\centering
\small
\caption{POS Probe Evaluation – Large Model, Layer 2}
\label{tab:pos_probe_layer_2}
\begin{tabular}{lccccc}
\toprule
\textbf{Label} & \textbf{Count} & \textbf{Accuracy} & \textbf{Precision} & \textbf{Recall} & \textbf{F1 Score} \\
\midrule
NOUN & 279984 & 1.000 & 1.000 & 1.000 & 1.000 \\
TRANSITIVE\_VERB & 151232 & 0.772 & 0.657 & 0.772 & 0.710 \\
INTRANSITIVE\_VERB & 59584 & 0.309 & 0.895 & 0.309 & 0.459 \\
COMMUNICATION\_VERB & 54352 & 0.476 & 1.000 & 0.476 & 0.645 \\
MOTION\_VERB & 68368 & 0.757 & 0.667 & 0.757 & 0.709 \\
CHANGE\_VERB & 14240 & 1.000 & 0.510 & 1.000 & 0.676 \\
ADJ & 59072 & 0.603 & 0.558 & 0.603 & 0.580 \\
LOCATION & 101984 & 0.654 & 0.829 & 0.654 & 0.731 \\
TEMP & 37664 & 0.367 & 0.535 & 0.367 & 0.436 \\
PREP & 115264 & 0.856 & 0.808 & 0.856 & 0.831 \\
RESULT & 68592 & 0.572 & 0.754 & 0.572 & 0.650 \\
CONJ & 51248 & 0.545 & 1.000 & 0.545 & 0.705 \\
\bottomrule
\end{tabular}
\end{table}

\begin{table}[H]
\centering
\small
\caption{Semantic Probe Evaluation – Large Model, Layer 0}
\label{tab:semantic_probe_layer_0}
\begin{tabular}{lccccc}
\toprule
\textbf{Label} & \textbf{Count} & \textbf{Accuracy} & \textbf{Precision} & \textbf{Recall} & \textbf{F1 Score} \\
\midrule
AGENT & 268576 & 1.000 & 0.959 & 1.000 & 0.979 \\
PATIENT & 151232 & 0.852 & 0.576 & 0.852 & 0.687 \\
ACTION & 279984 & 1.000 & 1.000 & 1.000 & 1.000 \\
LOCATION & 101984 & 0.315 & 0.858 & 0.315 & 0.461 \\
RELATION & 115264 & 0.387 & 0.823 & 0.387 & 0.526 \\
CONNECTOR & 51248 & 0.321 & 0.950 & 0.321 & 0.480 \\
RESULT & 68592 & 0.305 & 0.731 & 0.305 & 0.431 \\
OTHER & 153248 & 0.874 & 0.642 & 0.874 & 0.740 \\
\bottomrule
\end{tabular}
\end{table}
\begin{table}[H]
\centering
\caption{Semantic Probe Evaluation – Large Model, Layer 1}
\label{tab:semantic_probe_layer_1}
\begin{tabular}{lccccc}
\toprule
\textbf{Label} & \textbf{Count} & \textbf{Accuracy} & \textbf{Precision} & \textbf{Recall} & \textbf{F1 Score} \\
\midrule
AGENT & 268576 & 1.000 & 0.959 & 1.000 & 0.979 \\
PATIENT & 151232 & 0.785 & 0.652 & 0.785 & 0.713 \\
ACTION & 279984 & 1.000 & 1.000 & 1.000 & 1.000 \\
LOCATION & 101984 & 0.641 & 0.843 & 0.641 & 0.729 \\
RELATION & 115264 & 0.856 & 0.808 & 0.856 & 0.831 \\
CONNECTOR & 51248 & 0.574 & 0.944 & 0.574 & 0.714 \\
RESULT & 68592 & 0.546 & 0.771 & 0.546 & 0.639 \\
OTHER & 153248 & 0.893 & 0.818 & 0.893 & 0.854 \\
\bottomrule
\end{tabular}
\end{table}
\begin{table}[H]
\centering
\caption{Semantic Probe Evaluation – Large Model, Layer 2}
\label{tab:semantic_probe_layer_2}
\begin{tabular}{lccccc}
\toprule
\textbf{Label} & \textbf{Count} & \textbf{Accuracy} & \textbf{Precision} & \textbf{Recall} & \textbf{F1 Score} \\
\midrule
AGENT & 268576 & 1.000 & 0.959 & 1.000 & 0.979 \\
PATIENT & 151232 & 0.762 & 0.659 & 0.762 & 0.707 \\
ACTION & 279984 & 1.000 & 1.000 & 1.000 & 1.000 \\
LOCATION & 101984 & 0.641 & 0.843 & 0.641 & 0.729 \\
RELATION & 115264 & 0.856 & 0.808 & 0.856 & 0.831 \\
CONNECTOR & 51248 & 0.545 & 1.000 & 0.545 & 0.705 \\
RESULT & 68592 & 0.390 & 0.969 & 0.390 & 0.556 \\
OTHER & 153248 & 0.893 & 0.818 & 0.893 & 0.854 \\
\bottomrule
\end{tabular}
\end{table}

\subsection{Computational Resources and Software Environment}\label{sec:compute_requirements}
All the experiments were on an NVIDIA RTX A6000. We used PyTorch (v2.6.0), scikit-learn (v1.6.1), scipy (v1.12.0), seaborn (v0.13.2), Python (v3.9.21), matplotlib (v3.9.4), numpy (v1.26.4), and matplotlib (v3.9.4). Training the small models required around 21 minutes without tracking, and 116 with tracking of all metrics. For medium models, it takes around 28 minutes to train and 130 minutes to train with full tracking. For large models, it takes around 34 minutes to train and 139 minutes to train with full tracking. Training probes require approximately 9 minutes, 19 minutes, and 28 minutes for small, medium and large models, respectively.


\section{Mutual Information Estimation with MINE}\label{app:mine}

To quantify the flow and compression of information within transformer models, we estimate mutual information (MI) between input embeddings and internal layer representations. While several methods exist for MI estimation (e.g., k-nearest neighbours, contrastive approaches), we adopt Mutual Information Neural Estimation (MINE) \citep{pmlr-v80-belghazi18a} due to its scalability and effectiveness in high-dimensional settings.

In theory, a systematic decline in $I(X; Z_\ell)$ across layers and training steps should signal abstraction: the model progressively discards surface-level details while retaining task-relevant structure. This compression-based view of abstraction has been explored in other architectures \citep{8482098, elmoznino2024complexity}, and we examined whether similar dynamics emerge in transformer models during training.

Specifically, we tracked two MI quantities:  
(i) $I(X; Z_\ell)$ — the mutual information between the input embeddings and the hidden states at layer $\ell$, and  
(ii) $I(Z_\ell; Z_{\ell+1})$ — the mutual information between consecutive hidden layers.

Despite the theoretical appeal, our empirical findings (see Figures~\ref{fig:mi_trajectories}) showed that MI was highly variable across training steps and did not consistently align with the phase transitions identified via curvature or intrinsic dimensionality. These results suggest that MI, while informative in principle, may lack the temporal resolution and stability needed to serve as a primary diagnostic in \textsc{TRACE}.

We report full implementation details, training settings, and estimator architecture below.

\subsection{MINE Objective and Architecture}
MINE approximates the Donsker-Varadhan lower bound on mutual information using a neural critic function \( T_\theta : \mathcal{X} \times \mathcal{Z} \rightarrow \mathbb{R} \), parameterised by a multilayer perceptron (MLP). Given joint samples \( (x, z) \sim P_{XZ} \) and marginal samples formed by pairing \( x \sim P_X \) with independently sampled \( z' \sim P_Z \), the MI is estimated via:
\begin{equation}
\hat{I}_\theta(X; Z) = \mathbb{E}_{P_{XZ}}[T_\theta(x, z)] - \log \mathbb{E}_{P_X \otimes P_Z}[e^{T_\theta(x, z')}]
\end{equation}

Our implementation uses a 3-layer MLP with hidden dimensions \([128, 128, 1]\) and ReLU activations. MINE is used to estimating \( I(X; Z_\ell) \) — the MI between the input embeddings and layer \(\ell\) — it is also used to compute \( I(Z_\ell; Z_{\ell+1}) \), capturing how information is transmitted between adjacent layers. This allows us to analyse information bottlenecks, compression phases, and abstraction dynamics across the depth of the model.


\subsection{Training and Evaluation Protocol}
For each chosen training step $t$ of the model and its layers, we train a separate MINE estimator to convergence. The training protocol is as follows:

\begin{itemize}
    \item \textbf{Batch size}: 128 examples
    \item \textbf{Optimiser}: Adam optimiser \citep{kingma2014adam} with learning rate 0.001
    \item \textbf{Training steps}: 200 iterations
    \item \textbf{Positive samples}: Joint pairs \( (x, z) \) from the same forward pass
    \item \textbf{Negative samples}: \( (x, z') \) where \( z' \) is obtained by shuffling across the batch
\end{itemize}

\subsection{MI Across Layers and Models}
\begin{figure}[h]
    \centering
    \includegraphics[width=.48\linewidth]{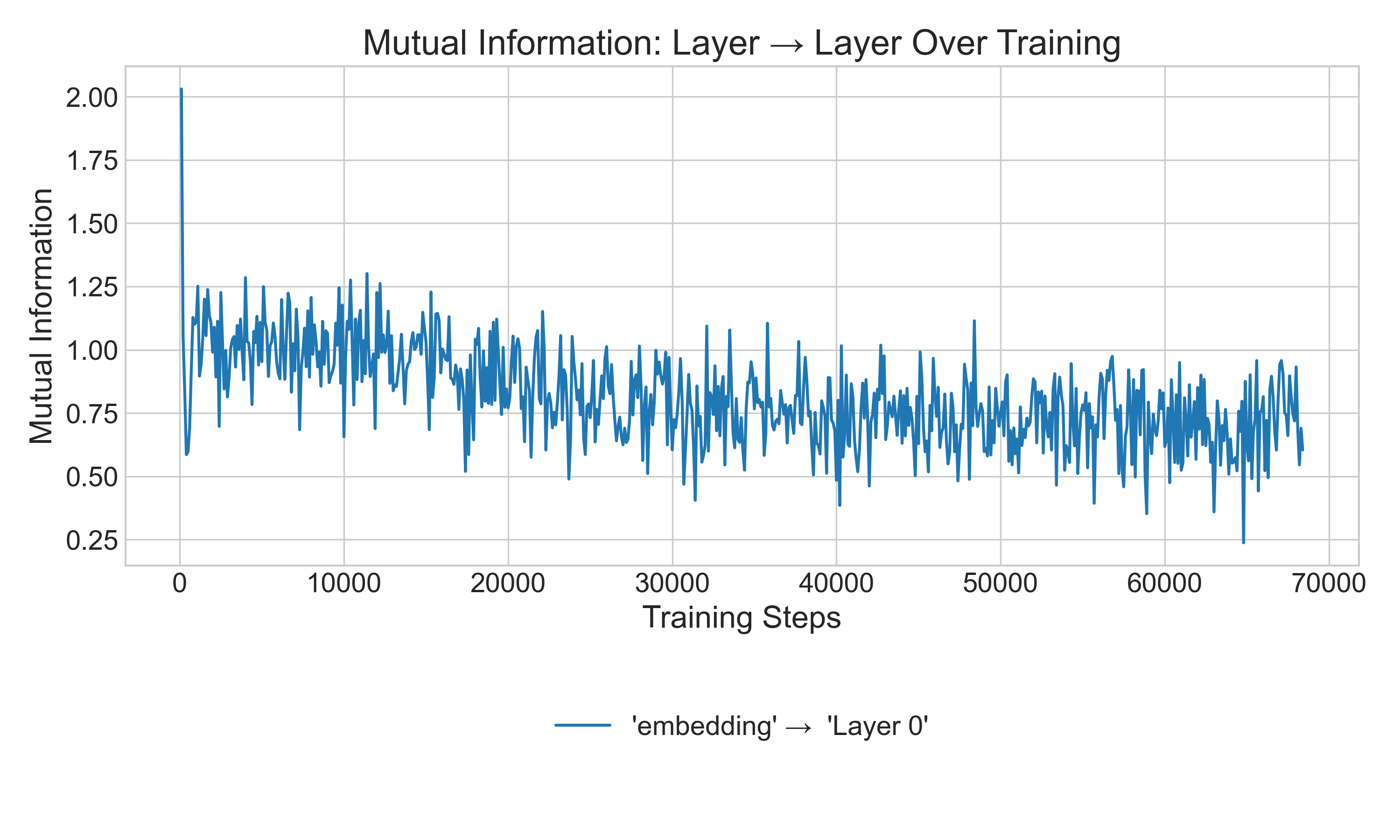}
    \includegraphics[width=.48\linewidth]{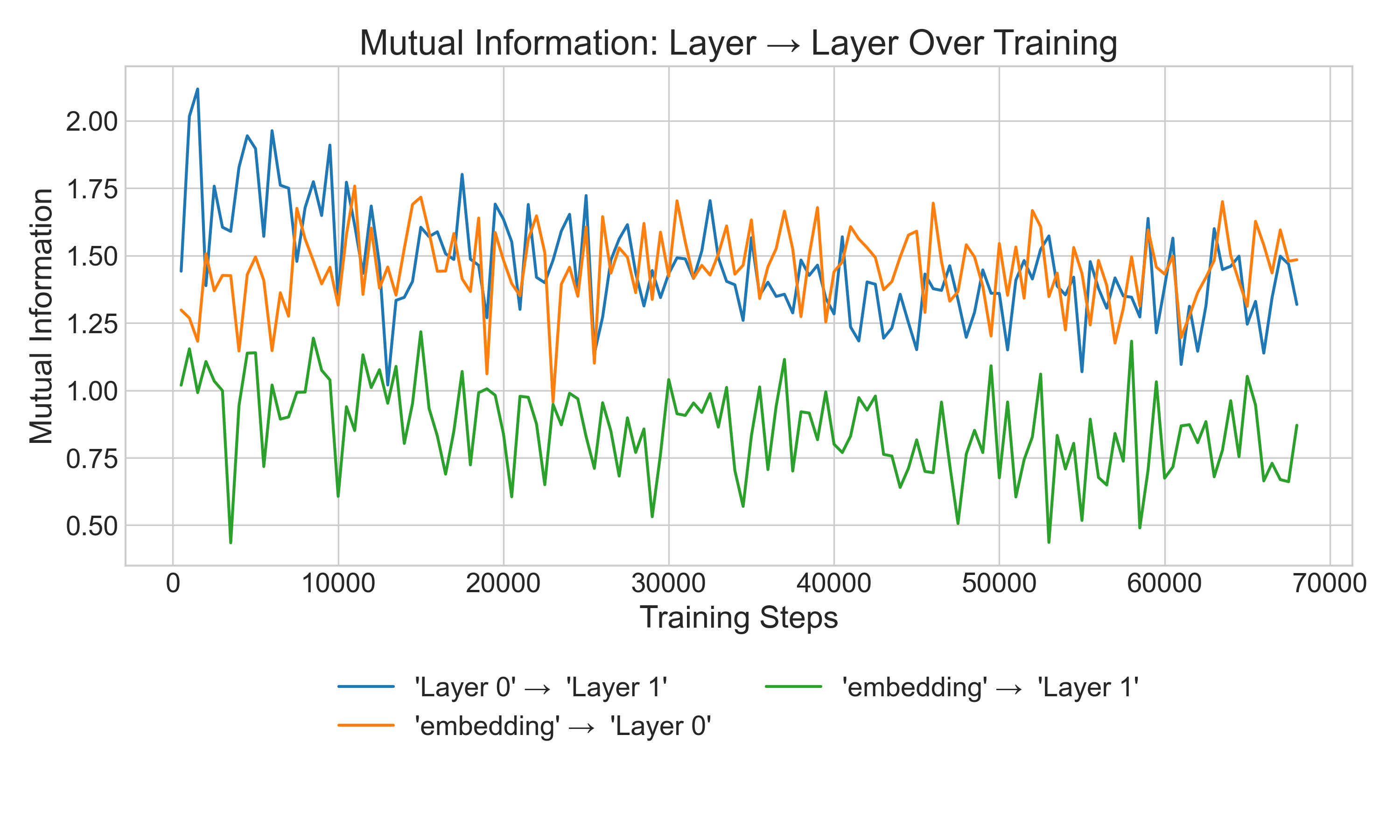}\\
    \includegraphics[width=.48\linewidth]{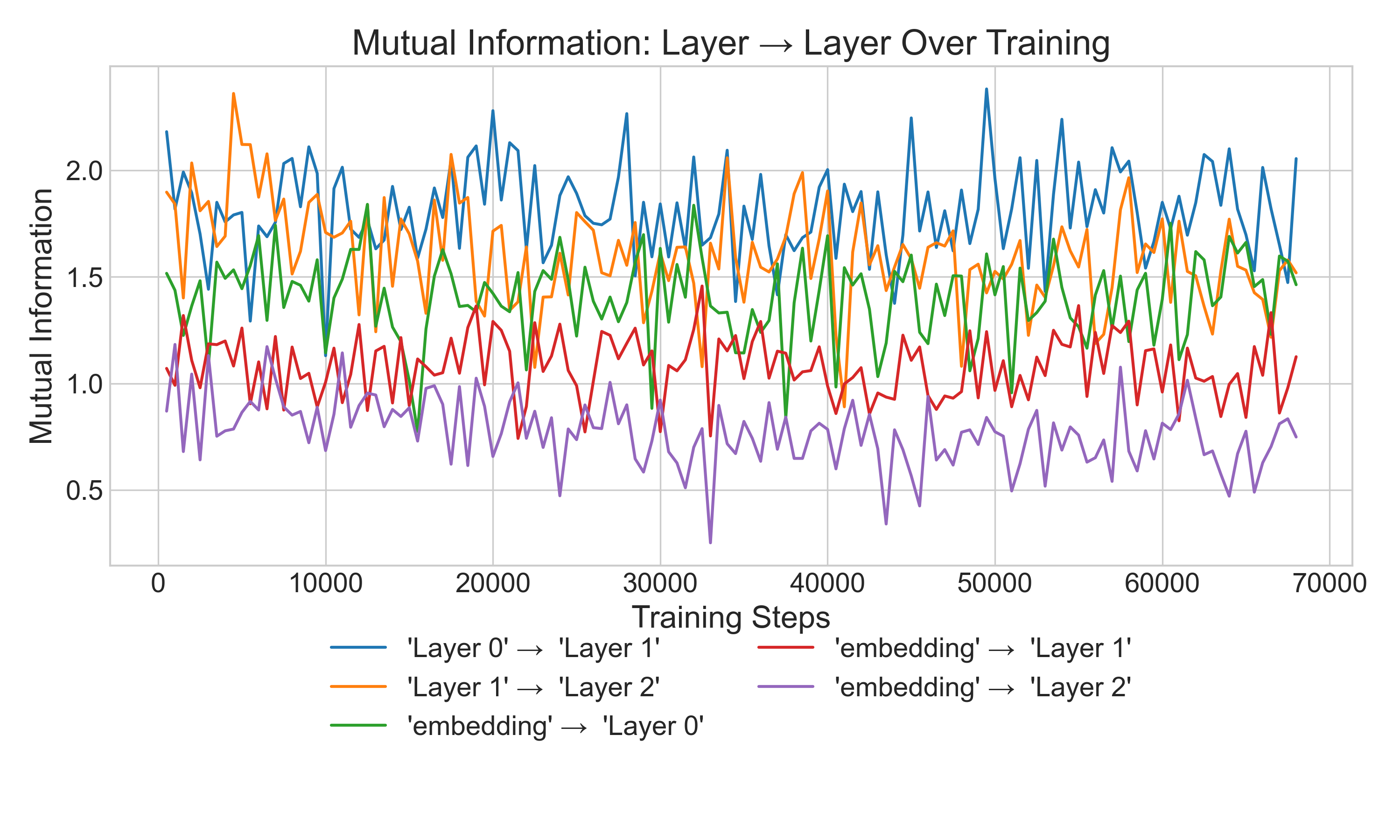}
    \caption{Mutual Information (MI) between adjacent layers over training steps for small (top-left), medium (top-right), and large (bottom) models. Each line represents the MI between one pair of layers (e.g., embedding $\rightarrow$ Layer 0, Layer 0 $\rightarrow$ Layer 1). Higher MI values suggest greater information flow or redundancy; drops indicate compression.}

    \label{fig:mi_trajectories}
\end{figure}

Figure~\ref{fig:mi_trajectories} shows how mutual information evolves during training across model scales. In small models (top-left), MI between the embedding and subsequent layers drops rapidly and stabilises early, suggesting early-stage compression and limited representational differentiation.

In medium models (top-right), we observe a pronounced dip in MI that aligns temporally with the phase transition identified in our diagnostics. This suggests that representational compression may act as a precursor or trigger for abstraction. Following this dip, MI values remain volatile across layers, reflecting a noisier or less stable reconfiguration of internal representations.

Large models (bottom) follow a broadly similar trend, though with key differences: the MI dip appears in most transitions, but is less evident or absent in others (Layer 1 $\rightarrow$ Layer 2). 

While the timing of these dips often aligns with the abstraction phase transition, the metric remains highly volatile. This instability suggests that mutual information, although partially correlated with representational restructuring, may be too noisy and inconsistent to serve as a reliable standalone indicator of abstraction onset.

\newpage



\end{document}